\RequirePackage[dvipsnames]{xcolor}
\PassOptionsToPackage{dvipsnames}{xcolor}

\documentclass{article}
\usepackage{amssymb}

\usepackage[final]{corl_2026} 

\hypersetup{
  pdftitle={CoPark: Learning Reactive Parking via Self-Play},
  pdfauthor={Jiarong Wei, Yanxing Chen, Sinuo Song, Yin Wu, Anna Rehr, Abhinav Valada}
}

\usepackage{amsfonts}
\usepackage{amsmath}
\usepackage{siunitx}
\usepackage{xfrac}
\usepackage{pifont}

\usepackage{tabularx}
\usepackage{booktabs}
\usepackage{makecell}
\usepackage[flushleft]{threeparttable}
\usepackage{multirow}
\usepackage{rotating}
\usepackage{microtype}

\usepackage{graphicx}
\usepackage{subcaption}
\usepackage{wrapfig}

\definecolor{darkgreen}{rgb}{0.0,0.5,0.0}
\usepackage[inline]{enumitem}
\usepackage{xspace}
\usepackage{caption}
\makeatletter
\renewenvironment{table}
  {\setlength{\abovecaptionskip}{\@nipsabovecaptionskip}%
   \setlength{\belowcaptionskip}{\@nipsbelowcaptionskip}%
   \@float{table}}
  {\end@float}
\makeatother
\usepackage{array}
\usepackage{float}
\usepackage[hang,flushmargin]{footmisc}
\usepackage[capitalize]{cleveref}

\crefname{section}{Sec.}{Secs.}
\Crefname{section}{Section}{Sections}
\Crefname{table}{Table}{Tables}
\crefname{table}{Tab.}{Tabs.}
\newcommand{\method}{\mbox{CoPark}\xspace}

\newlength{\mytablewidth}
\title{CoPark: Learning Reactive Parking via Self-Play}

\author{
  Jiarong Wei\textsuperscript{1,2}, 
  Yanxing Chen\textsuperscript{3}, 
  Sinuo Song\textsuperscript{3}, 
  Yin Wu\textsuperscript{2}, 
  Anna Rehr\textsuperscript{2}, 
  Abhinav Valada\textsuperscript{1} \\
  \textsuperscript{1}Department of Computer Science, University of Freiburg, Germany \\
  \textsuperscript{2}CARIAD SE, Germany \\
  \textsuperscript{3}Technical University of Munich, Germany
}

\begin{document}
\maketitle

\vspace{-7mm}
\begin{figure}[H]
\centering
\includegraphics[
  width=0.9\linewidth,
  trim=0cm 0.15cm 0cm 0.15cm
  ]{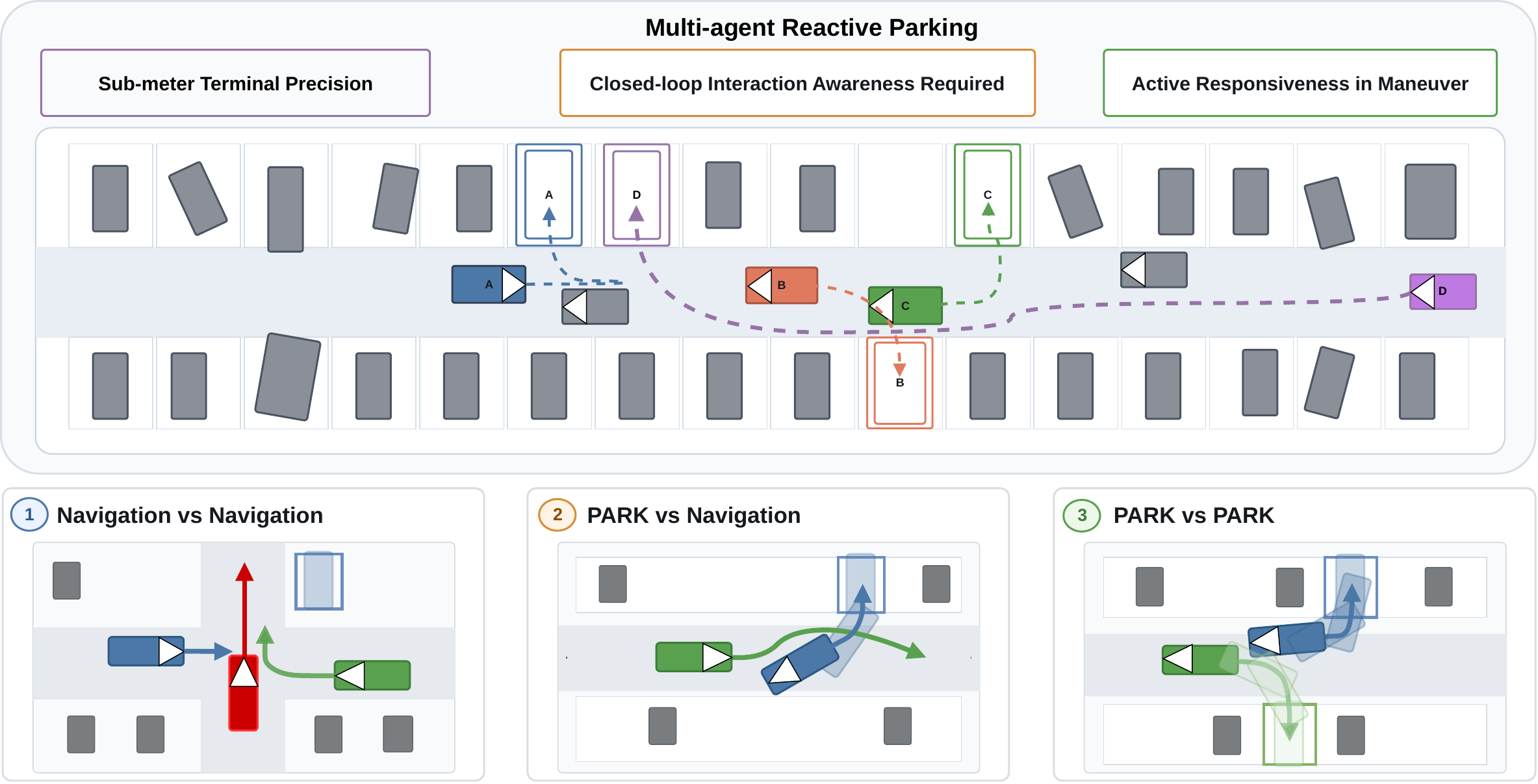}

\vspace{-1mm}

\caption{\small Reactive parking requires multiple vehicles in a shared lot to reach assigned slots from random initial poses, with strict terminal tolerance on slot pose, while remaining responsive to neighbors throughout the process. Bottom: interaction during navigation (1), parking maneuver (3), or both (2).}
\label{fig:teaser}

\vspace{-5mm}
\end{figure}

\vspace{0cm}
\begin{abstract}
Learning a single policy that reaches a goal with high geometric precision while interacting safely with nearby agents poses conflicting objectives. Precision favors commitment to a fixed geometric plan, whereas interaction requires immediate deviation when another agent intrudes, causing policies optimized for one objective to often fail at the other. We study this problem in the context of reactive autonomous parking, where multiple vehicles must reach assigned slots with sub-meter terminal accuracy while remaining responsive to neighboring vehicles throughout the maneuver. We propose \method, a self-play RL approach that learns reactive parking through multi-agent self-play, built on a residual-policy architecture. A precomputed offline plan provides a fixed action prior, while a residual head learns the reactive corrections.
This decomposition separates geometric consistency from interaction. The residual policy learns behaviors under self-play, where data and scripting fall short, while the fixed prior holds the slot-frame geometry that pure policies struggle to reach reliably.
The key design is a partner-threat-modulated, channel-asymmetric release of the prior. A continuous threat signal shifts authority of the longitudinal channel to the residual head to enable yielding, while the lateral channel remains anchored to the precomputed reference to preserve sub-meter slot alignment. A closed-loop refinement layer corrects residual terminal error from action-grid discretization.
We train our policy on six parking lots and evaluate zero-shot under a new closed-loop evaluation protocol built on Dragon Lake Parking (DLP) and DeepScenario Open 3D (DSC3D). \method achieves $\sim$70--85\% success with only 3--6\% collision rate, substantially outperforming classical, imitation-learning, and large-scale RL baselines. Importantly, the results demonstrate emergent interaction behaviors such as reverse-yielding, mid-maneuver yielding, tight-corridor passing, and queuing.\looseness=-1
\end{abstract}


\section{Introduction}
\label{sec:intro}

Self-play Reinforcement Learning (RL) has driven breakthroughs from board games and competitive control~\cite{samuel1959some,bansal2017emergent,silver2017master,silver2018general} to autonomous driving, where large-scale multi-agent simulation and closed-loop training yield the emergence of robust behavior on held-out scenarios~\cite{cusumano2025robust,kazemkhani2025gpudrive,cornelisse2025building,cornelisse2024human}. These successes are concentrated in settings shaped by strong lane priors and forward-only kinematics, where sub-meter terminal accuracy is not required. Tasks that simultaneously combine high-precision maneuvers, tight spatial constraints, and low-speed multi-agent interaction remain largely unaddressed by self-play.

Reactive parking is a representative instance (Fig.~\ref{fig:teaser}): each vehicle must navigate from a random feasible pose to its assigned slot with parking-grade accuracy while remaining responsive to neighbors throughout the maneuver. The task jointly demands sub-meter terminal precision under non-holonomic geometry, closed-loop interaction awareness in confined aisles, and active responsiveness during the parking maneuver rather than only during routing, a combination that existing paradigms address only in part.

Meeting these requirements together is hard since they pull against one another within a single policy. Robust interaction is difficult to learn from supervision: the scarce parking-interaction data resists imitation, and the combinatorial space of multi-agent encounters resists hand-scripting, which points to self-play as the paradigm for learning interactive behavior. Precision is the opposing difficulty. A policy learned purely from reward struggles to reach parking-grade accuracy, needing elaborate curricula even in the single-agent case~\cite{xu2025hrl, chai2022deep}, let alone while simultaneously learning to interact. Precision therefore calls for external geometric grounding beyond what reward alone discovers. This grounding, however, conflicts with reactivity: a fixed reference anchors the slot frame but, held rigidly, blocks the deviations that yielding requires. Reconciling precision and reactivity within one policy, without a behavior-specific controller for each interaction pattern, is the central challenge.

We resolve this challenge with \method, a self-play RL approach that learns reactive parking through multi-agent self-play. To supply precision without bottlenecking self-play, an inexpensive offline plan, a Stanley tracker over a Hybrid~A$^*$ path computed once per instance, enters the action logits as a fixed prior, sidestepping the unstable, curriculum-heavy precision learning that pure policies require. To keep this prior from obstructing interaction, a residual head modulates its authority by perceived partner threat and releases it anisotropically, relaxing the longitudinal channel so the policy can yield while anchoring the lateral channel to hold slot alignment. This instantiates a general principle: a prior should be released to match a task's anisotropic tolerance, anchored where tolerance is tight and relaxed where it is loose. A closed-loop refinement layer then continues acting after the plan ends, absorbing terminal error from action-grid discretization and partner-induced drift.

We train our policy on six parking lots and evaluate zero-shot under a new closed-loop evaluation protocol on DLP and DSC3D in PufferDrive~\cite{pufferdrive2025github}. The results demonstrate that \method achieves a $70$--$85\%$ success rate with a $3$--$6\%$ collision rate across both partner modes, substantially outperforming classical, imitation-learning, and pure large-scale RL baselines, while exhibiting emergent interaction behaviors such as reverse-yielding, mid-maneuver yielding, tight-corridor passing, and queuing.

To summarize, the main contributions are as follows:
\begin{enumerate}[topsep=0pt,itemsep=0pt]
    \item We formulate reactive parking, a closed-loop multi-agent task that jointly requires parking-grade terminal precision, full-process interaction awareness, and active responsiveness during the maneuver itself, a combination no prior approach addresses jointly.
    \item We propose \method, whose central mechanism is a partner-threat-modulated, channel-asymmetric release of an offline kinematic prior, combined with a closed-loop refinement layer for terminal precision.
    \item We design a closed-loop reactive-parking evaluation protocol that replays DLP and DSC3D scenes in the PufferDrive simulator, with reactive and non-reactive partner modes.
    \item We present extensive evaluations showing strong zero-shot performance across both datasets, substantially outperforming representative baselines in success rate, with emergent multi-agent behaviors.
\end{enumerate}
\section{Related Work}
\label{sec:related-work}

\textbf{Self-Play for Closed-Loop Autonomous Driving}: Self-play~\cite{samuel1959some,bansal2017emergent} has driven breakthroughs in zero-sum games~\cite{silver2017master,silver2018general}, and recent work transfers the idea to closed-loop autonomous driving. GigaFlow~\cite{cusumano2025robust} demonstrates that robust naturalistic driving behaviors emerge when self-play is scaled to long horizons with thousands of agents and per-agent reward conditioning. CaRL~\cite{jaeger2025carl}, not itself self-play, scales a PPO planner with deliberately simple rewards to match or surpass more elaborate recipes on standard nuPlan and CARLA benchmarks. GPUDrive~\cite{kazemkhani2025gpudrive} provides a million-FPS multi-agent simulator that makes such scale practical, and Cornelisse~\textit{et~al.}~\cite{cornelisse2025building} report near-closure of the train--test gap on held-out scenarios by scaling self-play to thousands of driving scenes. Follow-up work adds KL regularization toward a human reference policy~\cite{cornelisse2024human,chang2025spacer} to keep emergent driving styles human-compatible, or introduces asymmetric teacher--student dynamics~\cite{zhang2024learning} that broaden the training distribution toward rare but safety-critical situations. These methods operate on open road networks where lane structure constrains the solution space and centimeter-level terminal accuracy is never demanded. Reactive parking sits in the opposite regime, demanding sub-meter terminal precision, dense low-speed interaction in confined corridors, and online responsiveness throughout the parking maneuver, none of which these large-scale recipes are built to deliver.

\textbf{Prior-Guided and Residual Reinforcement Learning}: A complementary line of work injects external structure into RL to meet precision requirements that pure exploration struggles to satisfy. One branch uses demonstrations or reference policies as a soft signal~\cite{hester2018deep,schmalstieg2023learning,chandra2025diwa,nematollahi2022robot,schmalstieg2022learning}, encoding the teacher through auxiliary imitation losses or KL regularisation toward a behavioral-cloning baseline. A second branch is residual policy learning~\cite{Johannink2019residual,Silver2018residual,rana2020residual}, in which a learned policy refines a fixed sub-optimal controller rather than replacing it. The residual is typically applied in action space or as a log-prior on the policy distribution under a single global reliance scalar. In parking, HOPE~\cite{jiang2025hope} couples an RL agent with Reeds--Shepp curves via action masking, and DRIP~\cite{jiang2025drip} uses an RL-pretrained prior refined by a diffusion model. Most of these methods are trained in single-ego settings with static or scripted neighbors and treat the prior as a fixed structural component whose authority does not vary with the multi-agent state. Reactive parking instead requires the prior's authority to be modulated by partner threat and split across action channels, a form of structured prior release that prior-guided and residual RL have left unexplored.

\textbf{Multi-Agent Interaction Learning for Automated Parking}: Automated valet parking systems traditionally decouple multi-vehicle coordination from low-level motion. At the strategic layer, learning is applied to slot allocation~\cite{xie2023drl} and joint parking-and-charging scheduling~\cite{boateng2024automated}. At the motion layer, coordination is handled by infrastructure-coordinated predictive control with V2I~\cite{9091894} or by hierarchical path planning with multi-vehicle collision avoidance~\cite{shen2024cav}, both depending on explicit interaction models and centralized computation. Recent multi-agent RL approaches~\cite{tanner2022multi,chen2023conflict} jointly train multiple parking agents but focus on conflict-free planning under simplified kinematics rather than on closed-loop terminal precision. Beyond the hybrid prior-RL methods discussed above, learning-based parking is otherwise mostly single-agent~\cite{xu2025hrl,chai2022deep}, with a parallel line modeling multi-agent parking interaction for trajectory prediction~\cite{wei2025parkdiffusion,wei2026parkdiffusionpp} rather than for closed-loop control. To our knowledge, no prior work has learned interactive parking behavior across both the navigation and parking-maneuver phases under closed-loop multi-agent self-play.
\section{Problem Formulation}
\label{sec:problem-formulation}

We formulate reactive parking as a general-sum partially observable stochastic game~\cite{hansen2004dynamic} (POSG) $\langle \mathcal{N}, \mathcal{S}, \mathcal{A}, \mathcal{O}, P, R, \gamma \rangle$, where $N = |\mathcal{N}| \ge 2$ controlled vehicles share a parking lot, each pursuing its own assigned slot while remaining responsive to neighbors.

\textbf{Setting}: A parking lot is $\mathcal{L} = (\mathcal{D}, \mathcal{P}, \mathcal{O}_{\text{obs}}, \mathcal{G})$ with drivable region $\mathcal{D}$, slot poses $\mathcal{P} = \{p_m\}_{m=1}^M$, static obstacles $\mathcal{O}_{\text{obs}}$, and lane graph $\mathcal{G}$. Each vehicle $i$ has state $s_{i,t} = (x_{i,t}, y_{i,t}, \psi_{i,t}, v_{i,t})$ under bicycle kinematics, takes a discrete action $a_{i,t} = (a^{\text{acc}}_{i,t}, a^{\text{steer}}_{i,t})$ from an acceleration--steering grid, and receives a partial observation $o_{i,t}$ of ego state, the relative state of its $K$ nearest neighbors ($K{=}8$), local geometry, and target slot pose. Vehicles start at distinct feasible initial states $s_{i,0}$ on $\mathcal{D}$ and are assigned target slots via an injective map $\sigma : \mathcal{N} \to \{1, \dots, M\}$. Each trajectory decomposes into navigation and parking-maneuver phases.

\textbf{Success and termination}: In the slot-aligned frame of $p_{\sigma(i)}$, agent $i$ succeeds when
\begin{equation}
\label{eq:success}
\|e_{i,t}^{\text{slot}}\|_2 \le \Delta_d, \quad \bigl|\mathrm{wrap}_{\pi}(\psi_{i,t} - \psi_{\sigma(i)})\bigr| \le \Delta_\psi, \quad |v_{i,t}| \le \Delta_v,
\end{equation}
holds for $K_s$ consecutive steps. The heading is wrapped modulo $\pi$, accepting both forward-in and reverse-in orientations. Trajectories also terminate on collision (vehicle-vehicle or vehicle-static) or timeout. The per-agent reward $R(s_{i,t}, a_{i,t}, S_t)$, with joint state $S_t = \{s_{j,t}\}_{j \in \mathcal{N}}$, couples agents through terminal events and generic dense safety terms, and contains no behavior-specific rewards for yielding, queuing, or reverse-yielding.

\textbf{Objective}: We restrict to homogeneous parameter-sharing policies $\pi_\theta : \mathcal{O} \to \Delta(\mathcal{A})$ with learning objective
\begin{equation}
\label{eq:objective}
\theta^\star = \arg\max_\theta \;\mathbb{E}_{(\mathcal{L}, \{s_{i,0}\}, \sigma) \sim \mathcal{Q}} \,\mathbb{E}_{\pi_\theta}\!\left[\sum_{i \in \mathcal{N}} \sum_{t=0}^{T} \gamma^{t} R(s_{i,t}, a_{i,t}, S_t)\right],
\end{equation}
where $\mathcal{Q}$ is the joint task distribution over parking lots, initial states, and slot assignments, and $T$ is the episode horizon, with each agent's reward set to zero after its own termination. During training, every partner in $S_t$ is itself controlled by $\pi_\theta$, making the optimization a multi-agent self-play process. The learned policy is ego-centric and can be deployed against arbitrary partners that need not share it (Sec.~\ref{sec:benchmark}).
\begin{figure}[!t]
\centering
\vspace{-4pt}
\includegraphics[
  width=1.00\linewidth,
]{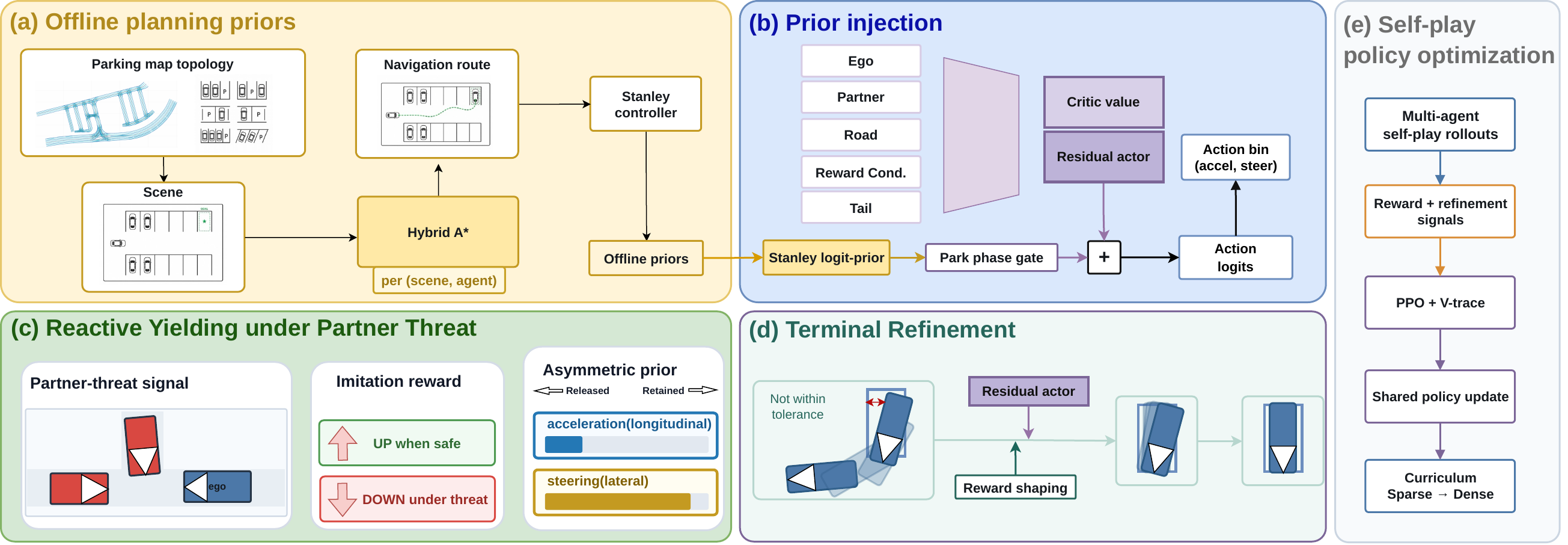}
\vspace{-4pt}
\caption{\small Architecture of \method. (a) Offline planning priors: Hybrid~A$^*$ plan and Stanley tracker, precomputed per (scene, agent), supply a geometric reference. (b) Prior injection: a residual actor over ego, partner, road, reward-conditioning, and tail blocks adds logits to the phase-gated Stanley logit-prior. (c) Reactive yielding: the partner-threat signal modulates imitation reward and asymmetrically releases the longitudinal prior while anchoring the lateral channel. (d) Terminal refinement: the residual actor continues acting beyond plan completion under a refinement reward toward the slot frame. (e) Self-play policy optimization: multi-agent rollouts feed PPO with V-trace under a sparse-to-dense scene-density curriculum.}
\label{fig:architecture}
\vspace{-6pt}
\end{figure}
\section{Method}
\label{sec:method}
In this section, we introduce \method's four key components (Fig.~\ref{fig:architecture}). The offline kinematic prior (Sec.~\ref{sec:method-overview}) and its residual injection (Sec.~\ref{sec:method-policy}) provide terminal precision. The reactive yielding under partner threat (Sec.~\ref{sec:method-park}) keeps the policy responsive to neighbors throughout the parking maneuver, and the closed-loop refinement layer (Sec.~\ref{sec:method-refine}) corrects residual terminal error. All four are trained jointly under multi-agent self-play with a scene-density curriculum (Sec.~\ref{sec:method-curriculum}).

\subsection{Offline Kinematic Prior}
\label{sec:method-overview}

We precompute a fixed kinematic teacher per task instance to bootstrap the policy beyond pure self-play's budget. For every $(\mathcal{L}, i)$ pair, a Hybrid~A$^*$~\cite{Dolgov2010PathPF} search with a Reeds--Shepp~\cite{Reeds1990OPTIMALPF} analytic shot under static-only collision checks, leaving dynamic interaction to the residual, yields a navigation route on $\mathcal{G}$ from $s_{i,0}$ to a preparation pose $p_i^{\text{prep}}$, sampled at uniform arc length into subgoals $\{g_{i,1}, \ldots, g_{i,K_i}\}$, together with a maneuvering trajectory $\tau_i$ from $p_i^{\text{prep}}$ to $p_{\sigma(i)}$ on the execution action grid. The subgoals shape navigation progress and $\tau_i$ supplies the reference modulated in Sec.~\ref{sec:method-park}. The agent transitions one-way from the navigation to the maneuver phase upon entering a tolerance neighborhood of $p_i^{\text{prep}}$, defining the phase indicator $\phi_{i,t}$ used below.

\subsection{Residual Policy with Prior Injection}
\label{sec:method-policy}

A single parameter-sharing categorical policy $\pi_\theta(a \mid s)$, where $s$ abbreviates the ego observation $o_{i,t}$ of Sec.~\ref{sec:problem-formulation}, acts at every step of both phases, with the offline prior engaged only during the parking maneuver. The teacher trajectory $\tau_i$ is read by a Stanley controller~\cite{hoffmann2007stanley} which emits a target command $(\hat a_{i,t}, \hat \delta_{i,t})$, encoded as a pair of per-channel Gaussian log-priors $\log p_{\text{prior}}^{\text{acc}}$ and $\log p_{\text{prior}}^{\text{steer}}$ over the $7 \times 13$ acceleration--steering action grid and added to the policy logits:
{\small
\begin{equation}
\label{eq:policy}
\log \pi_\theta(a \mid s) \;\propto\; z_\theta(s, a) \;+\; \mathbf{1}[\phi_{i,t} = \mathrm{Maneuver}]\bigl(\alpha^{\text{acc}}(s)\,\log p_{\text{prior}}^{\text{acc}}(a) \,+\, \alpha^{\text{steer}}(s)\,\log p_{\text{prior}}^{\text{steer}}(a)\bigr),
\end{equation}
}
with $z_\theta(s, a)$ a bounded residual logit head and $\alpha^{\text{acc}}(s), \alpha^{\text{steer}}(s)$ channel-wise reliance scalars derived from a learned base $\alpha(s) \in [\alpha_{\min}, \alpha_{\max}]$. The Stanley controller never writes to the simulator; the reference enters only as a structural pull on the policy distribution, whose dynamics under partner threat are defined in Sec.~\ref{sec:method-park}.

\subsection{Reactive Yielding under Partner Threat}
\label{sec:method-park}

The prior pull of Eq.~\ref{eq:policy} must relax in proportion to partner activity, asymmetrically across the two action channels, and continuously rather than in switched modes. 

\textbf{Partner-threat signal}: We summarize the instantaneous interaction state of agent $i$ by
\begin{equation}
\label{eq:threat}
\text{threat}_i \;=\; \sum_{j \in \mathcal{N}_K(i)}\, [-\dot d_{ij}]_{+}\,\exp\!\bigl(-d_{ij}/d_{\text{decay}}\bigr),
\end{equation}
aggregating per-partner contributions over the $K$ nearest neighbors $\mathcal{N}_K(i)$ that form the partner block of the observation (Sec.~\ref{sec:problem-formulation}). Each contribution combines the rectified closing rate $[-\dot d_{ij}]_{+} = \max(0, -\dot d_{ij})$, where $d_{ij} = \|p_i - p_j\|_2$ is the ego--partner distance and $\dot d_{ij}$ its time derivative, with an exponentially decaying range envelope. The signal enters the per-step reward as a dense closing penalty $-\beta_{\text{dyn}}\,\text{threat}_i$, and is mapped into a saturated interaction intensity through a smooth squash,
\[
\rho(s) \;=\; 1 - \exp\!\bigl(-\text{threat}_i/\tau_\rho\bigr) \;\in\; [0, 1],
\]
rising continuously from $0$ in the empty lot toward $1$ under sustained dense interaction.

\textbf{Threat-modulated imitation reward}: The Stanley command also enters as a behavior-cloning reward, with weight relaxed continuously by $\rho(s)$:
\begin{equation}
\label{eq:imitation}
r^{\text{imit}}_{i,t} \;=\; \alpha_{\text{imit}}\,\bigl(1 - \rho(s)\bigr)\,w_{\text{trust}}(s)\,\exp\!\bigl(-\|u_{i,t} - \hat u_{i,t}\|_2 / \sigma_{\text{imit}}\bigr),
\end{equation}
where $u_{i,t}, \hat u_{i,t} \in [-1,1]^2$ are the per-channel-normalized policy and teacher commands and $w_{\text{trust}}(s) \in [0,1]$ is the multiplicative trust-weight
\begin{equation}
\label{eq:trust}
w_{\text{trust}}(s) \;=\; \underbrace{\exp\bigl(-\|e_{\text{ct}}(s)\|/\sigma_{\text{ct}}\bigr)}_{\text{cross-track suppression}} \;\cdot\; \underbrace{\sigma\!\bigl(\beta_{\text{stat}}(d_{\text{stat}}(s) - d_0)\bigr)}_{\text{static-clearance gate}} \;\cdot\; \underbrace{\max(0,\, 1 - \xi(s))}_{\text{plan-progress decay}},
\end{equation}
combining teacher cross-track error $e_{\text{ct}}(s)$, distance to the nearest static edge $d_{\text{stat}}(s)$, and the teacher's plan-progress fraction $\xi(s) \in [0, 1]$ ($\sigma$ is the logistic sigmoid). The imitation signal therefore never penalizes legitimate safety deviations and vanishes once the teacher has finished its plan.

\textbf{Asymmetric channel-wise prior modulation}: The reliance scalars of Eq.~\ref{eq:policy} contract along the same $\rho(s)$ but at different rates per channel,
\begin{equation}
\label{eq:split-reliance}
\alpha^{c}(s) \;=\; \alpha(s)\,\bigl(1 - \kappa^c\,\rho(s)\bigr), \qquad c \in \{\text{acc}, \text{steer}\},
\end{equation}
with $\kappa^{\text{acc}} \gg \kappa^{\text{steer}}$ as the only design asymmetry. The longitudinal prior relaxes toward $\alpha(s)(1 - \kappa^{\text{acc}})$, handing authority to the residual to brake, hold, or reverse, while the lateral prior persists near $\alpha(s)(1 - \kappa^{\text{steer}})$, anchoring the slot frame. Sec.~\ref{sec:ablations} (Table~\ref{tab:asym-ablation}) shows that the channel asymmetry $\kappa^{\text{acc}} - \kappa^{\text{steer}}$, not the release magnitude, is the operative design choice.

\subsection{Closed-Loop Terminal Refinement}
\label{sec:method-refine}

A single rollout of $\pi_\theta$ through $\tau_i$ rarely lands within the success tolerance of Eq.~\ref{eq:success} due to action-grid discretization and yielding-induced drift. We let the same $\pi_\theta$ continue acting beyond plan completion; once arc progress along $\tau_i$ stalls and the agent remains outside tolerance, the per-step reward switches to
\begin{equation}
\label{eq:refine-reward}
r^{\text{ref}}_{i,t} \;=\; \beta_{\text{red}}\,\eta\,\tanh\!\bigl((\|e_{i,t-1}^{\text{slot}}\|_2 - \|e_{i,t}^{\text{slot}}\|_2)/\eta\bigr) \;+\; \beta_{\text{hold}}\,\mathbf{1}\bigl[e_{i,t}^{\text{slot}} \in \mathcal{E}_{\text{tol}} \,\wedge\, |v_{i,t}| \le \Delta_v\bigr],
\end{equation}
where $\mathcal{E}_{\text{tol}}$ is the pose tolerance set from Eq.~\ref{eq:success} and $\eta$ is the tanh saturation scale. The first term is a smoothly saturating directed gradient toward the slot, bounded by $\pm\beta_{\text{red}}\eta$. The second is a hold bonus toward the $K_s$-step consecutive hold of the success criterion. Refinement ends in success when Eq.~\ref{eq:success} holds for $K_s$ steps and in failure when the budget $K^{\max}_{\text{ref}}$ is exhausted. The imitation reward of Eq.~\ref{eq:imitation} is naturally suppressed inside refinement since the teacher's plan has terminated.

\subsection{Self-Play Policy Optimization}
\label{sec:method-curriculum}

We optimize $\pi_\theta$ under multi-agent self-play with per-agent reward conditioning following GigaFlow's~\cite{cusumano2025robust} recipe. Training uses PPO~\cite{schulman2017proximal} with V-trace~\cite{espeholt2018impala} advantages through PufferLib~\cite{suarez2024pufferlib}. A single parameter set $\theta$ is shared across phases, agents, and episodes. Scenes are sampled under a sparse-to-dense scene-density curriculum set at data-preparation time (Appendix~\ref{app:scene-prep}). The phase information enters only through $\phi_{i,t}$ in $o_{i,t}$ and the phase-conditional prior of Eq.~\ref{eq:policy}. Completed agents are removed from the rollout on termination.
\section{Reactive Parking Evaluation Protocol}
\label{sec:benchmark}
For zero-shot evaluation of policies trained on custom layouts, we integrate Dragon Lake Parking (DLP)~\cite{DLP} and DeepScenario Open 3D (DSC3D)~\cite{DSC3D} into the PufferDrive~\cite{pufferdrive2025github} simulator under Sec.~\ref{sec:problem-formulation}, extracting every source-data parking trajectory as one ego-episode. This turns real logged parking trajectories, which carry no interaction protocol of their own, into the first closed-loop, reactive evaluation setting for multi-agent parking. Following GigaFlow's~\cite{cusumano2025robust} closed-loop protocol, each method controls a single ego while non-ego partners run in one of two scripted modes: reactive (IDM~\cite{treiber2000idm} along logged routes, decelerating in response to the ego) and non-reactive (log replay regardless of the ego), a self-play partners mode is reported for \method in Appendix~\ref{app:self-play-eval}. Each episode is scored on seven metrics: SR, Coll, and Off as ranking metrics over all episodes; PErr and HErr as terminal-precision metrics over each method's successful subset, and Path and Manv as diagnostic metrics read jointly with SR and Coll. Tolerances, the $91$-way action interface, and other implementation details are in Appendix~\ref{app:benchmark}.
\section{Experimental Results}
\label{sec:experiments}

In this section, we empirically validate \method under the reactive parking evaluation protocol of Sec.~\ref{sec:benchmark}, characterize its emergent multi-agent behaviors, and ablate each design choice.

\subsection{Training Setup and Baselines}
\label{sec:setup}

\textbf{Training Setup}: We train \method under multi-agent self-play on \num{2} NVIDIA RTX~A6000 GPUs for \num{4e9} environment steps, sustaining $\num{120}$k simulator steps per second at episode length \num{400}, using six custom parking-lot layouts pre-generated into \num{1024} scenes with a three-tier scene-density curriculum ($0.25/0.5/0.75$ occupancy). More details are presented in Appendix~\ref{app:scene-prep}. We evaluate zero-shot under the evaluation protocol of Sec.~\ref{sec:benchmark} under both partner modes.

\textbf{Baselines}: Seven baselines span three classical planners (RS~\cite{Reeds1990OPTIMALPF}, Hybrid~A$^*$~\cite{Dolgov2010PathPF}, MA-MPC~\cite{9091894}), an interaction-aware parking RL method (HOPE~\cite{jiang2025hope}), two large-scale RL recipes (GigaFlow~\cite{cusumano2025robust}, CaRL~\cite{jaeger2025carl}), and an imitation-learning planner (Diffusion Planner~\cite{zheng2025diffusion}, marked $^\ddagger$ because it is trained directly on DLP/DSC3D demonstrations and is therefore not zero-shot). Adaptation details are presented in Appendix~\ref{app:baselines}.

\subsection{Main Results}
\label{sec:main-results}

\begin{table}[!t]
\setlength{\abovecaptionskip}{6pt}
\vspace{6pt}
\centering
\small
\setlength{\tabcolsep}{3pt}
\renewcommand{\arraystretch}{0.9}
\resizebox{\linewidth}{!}{%
\begin{tabular}{c|l|ccccccc|ccccccc}
\toprule
\multicolumn{1}{c|}{\multirow{2}{*}[-0.5ex]{\textbf{Partner}}}
& \multicolumn{1}{c|}{\multirow{2}{*}[-0.5ex]{\textbf{Method}}}
& \multicolumn{7}{c|}{\textbf{DLP}~\cite{DLP}}
& \multicolumn{7}{c}{\textbf{DSC3D}~\cite{DSC3D}} \\
\cmidrule{3-9}\cmidrule{10-16}
& & SR\,[\%] & Path\,[m] & Manv & PErr\,[m] & HErr\,[$^\circ$] & Coll\,[\%] & Off\,[\%]
& SR\,[\%] & Path\,[m] & Manv & PErr\,[m] & HErr\,[$^\circ$] & Coll\,[\%] & Off\,[\%] \\
\midrule
\multirow{8}{*}{\shortstack{Reactive\\Agents}}
& RS~\cite{Reeds1990OPTIMALPF}                              & $5.8$  & $17.9$ & $1.7$ & $\mathbf{0.1}$ & $\mathbf{0.8}$ & $64.9$ & $5.6$ & $7.9$  & $11.2$ & $0.5$ & $\mathbf{0.1}$ & $\mathbf{0.6}$ & $64.8$ & $6.4$ \\
& Hybrid A$^{*}$~\cite{Dolgov2010PathPF}                    & $11.2$ & $22.4$ & $0.9$ & $0.1$ & $1.0$ & $57.6$ & $1.2$ & $6.9$ & $16.4$ & $0.6$ & $0.1$ & $0.8$ & $59.7$ & $1.8$ \\
& MA-MPC~\cite{9091894}                                     & $58.5$ & $35.2$ & $1.3$ & $0.2$ & $1.5$ & $10.2$ & $0.5$ & $46.8$ & $24.2$ & $1.0$ & $0.2$ & $1.7$ & $13.0$ & $0.8$ \\
& HOPE$^{\dagger}$~\cite{jiang2025hope}                     & $24.8$ & $41.5$ & $8.4$ & $0.2$ & $2.2$ & $38.2$ & $1.5$ & $8.4$  & $26.5$ & $5.9$ & $0.3$ & $3.8$ & $52.6$ & $2.5$ \\
& GigaFlow~\cite{cusumano2025robust}                        & $50.8_{\pm 1.7}$ & $43.8_{\pm 1.4}$          & $1.1_{\pm 0.1}$ & $0.5_{\pm 0.1}$ & $5.5_{\pm 0.5}$ & $\mathbf{3.8}_{\pm 0.7}$  & $0.8_{\pm 0.2}$ & $45.5_{\pm 1.6}$ & $27.5_{\pm 1.0}$ & $1.2_{\pm 0.1}$ & $0.5_{\pm 0.1}$ & $5.8_{\pm 0.5}$ & $\mathbf{3.3}_{\pm 0.8}$  & $1.2_{\pm 0.3}$ \\
& CaRL~\cite{jaeger2025carl}                                 & $45.0_{\pm 1.7}$ & $42.5_{\pm 1.4}$          & $1.2_{\pm 0.1}$ & $0.5_{\pm 0.1}$ & $6.0_{\pm 0.5}$ & $4.5_{\pm 0.7}$  & $1.0_{\pm 0.2}$ & $40.0_{\pm 1.6}$ & $26.4_{\pm 1.0}$ & $1.3_{\pm 0.1}$ & $0.5_{\pm 0.1}$ & $6.2_{\pm 0.5}$ & $4.0_{\pm 0.8}$  & $1.5_{\pm 0.3}$ \\
& Diffusion Planner$^{\ddagger}$~\cite{zheng2025diffusion}  & $54.6_{\pm 1.5}$ & $25.8_{\pm 0.8}$          & $1.7_{\pm 0.1}$ & $0.3_{\pm 0.0}$ & $2.5_{\pm 0.2}$ & $10.5_{\pm 1.0}$ & $0.5_{\pm 0.2}$ & $48.9_{\pm 1.6}$ & $19.8_{\pm 1.0}$ & $1.7_{\pm 0.1}$ & $0.3_{\pm 0.0}$ & $2.5_{\pm 0.2}$ & $13.2_{\pm 1.2}$ & $0.8_{\pm 0.3}$ \\
& \textbf{\method (Ours)}                                   & $\mathbf{84.7}_{\pm 1.4}$ & $47.6_{\pm 0.6}$ & $1.7_{\pm 0.2}$ & $0.4_{\pm 0.0}$ & $3.9_{\pm 0.2}$ & $4.1_{\pm 0.7}$ & $\mathbf{0.0}_{\pm 0.0}$ & $\mathbf{75.8}_{\pm 1.7}$ & $29.6_{\pm 0.5}$ & $1.7_{\pm 0.2}$ & $0.4_{\pm 0.0}$ & $3.3_{\pm 0.2}$ & $3.5_{\pm 0.8}$ & $\mathbf{0.0}_{\pm 0.0}$ \\
\midrule
\multirow{8}{*}{\shortstack{Non-Reactive\\Agents}}
& RS~\cite{Reeds1990OPTIMALPF}                              & $3.9$  & $18.0$ & $1.7$ & $\mathbf{0.1}$ & $\mathbf{1.0}$ & $77.8$ & $6.2$ & $5.5$  & $11.3$ & $0.5$ & $\mathbf{0.1}$ & $\mathbf{0.7}$ & $71.7$ & $6.9$ \\
& Hybrid A$^{*}$~\cite{Dolgov2010PathPF}                    & $8.0$  & $22.6$ & $0.9$ & $0.1$ & $1.2$ & $66.7$ & $1.5$ & $4.9$ & $16.5$ & $0.6$ & $0.1$ & $0.9$ & $65.7$ & $2.1$ \\
& MA-MPC~\cite{9091894}                                     & $49.0$ & $34.6$ & $1.3$ & $0.2$ & $1.6$ & $16.5$ & $0.7$ & $38.5$ & $23.7$ & $1.0$ & $0.2$ & $1.8$ & $19.5$ & $1.1$ \\
& HOPE$^{\dagger}$~\cite{jiang2025hope}                     & $19.8$ & $41.7$ & $8.4$ & $0.2$ & $2.5$ & $45.1$ & $1.8$ & $6.5$  & $26.6$ & $6.0$ & $0.3$ & $4.2$ & $58.3$ & $3.0$ \\
& GigaFlow~\cite{cusumano2025robust}                        & $47.8_{\pm 1.7}$ & $44.0_{\pm 1.4}$          & $1.1_{\pm 0.1}$ & $0.5_{\pm 0.1}$ & $5.8_{\pm 0.5}$ & $\mathbf{5.2}_{\pm 0.8}$  & $1.1_{\pm 0.3}$ & $41.9_{\pm 1.6}$ & $27.7_{\pm 1.0}$ & $1.2_{\pm 0.1}$ & $0.5_{\pm 0.1}$ & $6.0_{\pm 0.5}$ & $\mathbf{5.5}_{\pm 0.9}$  & $1.5_{\pm 0.3}$ \\
& CaRL~\cite{jaeger2025carl}                                 & $42.0_{\pm 1.7}$ & $42.7_{\pm 1.4}$          & $1.2_{\pm 0.1}$ & $0.5_{\pm 0.1}$ & $6.3_{\pm 0.5}$ & $6.2_{\pm 0.8}$  & $1.3_{\pm 0.3}$ & $36.5_{\pm 1.6}$ & $26.6_{\pm 1.0}$ & $1.3_{\pm 0.1}$ & $0.5_{\pm 0.1}$ & $6.5_{\pm 0.5}$ & $6.3_{\pm 0.9}$  & $1.8_{\pm 0.3}$ \\
& Diffusion Planner$^{\ddagger}$~\cite{zheng2025diffusion}  & $60.3_{\pm 1.5}$ & $26.0_{\pm 0.8}$          & $1.7_{\pm 0.1}$ & $0.3_{\pm 0.0}$ & $2.7_{\pm 0.2}$ & $5.8_{\pm 0.8}$ & $0.6_{\pm 0.2}$ & $53.7_{\pm 1.6}$ & $20.0_{\pm 1.0}$ & $1.7_{\pm 0.1}$ & $0.3_{\pm 0.0}$ & $2.7_{\pm 0.2}$ & $7.4_{\pm 1.0}$ & $1.0_{\pm 0.3}$ \\
& \textbf{\method (Ours)}                                   & $\mathbf{79.7}_{\pm 1.6}$ & $47.7_{\pm 0.6}$ & $1.8_{\pm 0.2}$ & $0.4_{\pm 0.0}$ & $4.0_{\pm 0.2}$ & $6.0_{\pm 0.8}$ & $\mathbf{0.0}_{\pm 0.0}$ & $\mathbf{69.8}_{\pm 1.8}$ & $29.7_{\pm 0.5}$ & $1.8_{\pm 0.2}$ & $0.4_{\pm 0.0}$ & $3.5_{\pm 0.2}$ & $5.8_{\pm 1.0}$ & $\mathbf{0.0}_{\pm 0.0}$ \\
\bottomrule
\end{tabular}%
}
\caption{\small Zero-shot evaluation on DLP and DSC3D under both partner reaction modes ($\pm$\,std over $5$ seeds; classical planners are deterministic). \method is trained on the custom layouts only, with no fine-tuning. Bold marks the best per column within each partner mode for SR, PErr, HErr, Coll, and Off. Path and Manv are diagnostic rather than ranked. $^{\ddagger}$Diffusion Planner is trained directly on DLP and DSC3D human demonstrations, so it is in-distribution rather than zero-shot. $^{\dagger}$HOPE uses a released checkpoint (no training seeds; not strictly zero-shot on DLP).}
\label{tab:main-results}
\end{table}

\textbf{Zero-shot Evaluation Results}: \method leads on SR (\SI{84.7}{\percent} / \SI{79.7}{\percent} / \SI{75.8}{\percent} / \SI{69.8}{\percent}) across two datasets under both partner reaction modes. Diffusion Planner ($\sim$\SIrange{49}{60}{\percent} in-distribution) and the large-scale RL recipes GigaFlow / CaRL ($\sim$\SIrange{37}{51}{\percent}) follow. GigaFlow leads on collisions (\SI{3.8}{\percent} on DLP-Reactive, \method second at \SI{4.1}{\percent}), and \method is the only method reaching \SI{0}{\percent} off-road in all conditions. Classical planners lead on terminal precision (\SI{0.1}{\meter} PErr) via analytic geometry. Learning-based methods sit at \SIrange{0.2}{0.5}{\meter}. Manv separates the large-scale RL methods: \method's $\sim$\num{1.7} maneuvers per episode match the human demonstration distribution (Appendix~\ref{app:path-manv-dist}) via reverse-in and re-tries, while GigaFlow / CaRL plateau on slots reachable in a single drive-in (Manv \numrange{1.1}{1.3}). \looseness=-1

\textbf{Failure Modes}: Open-loop classical planners (RS, Hybrid~A$^*$+Stanley) lack partner awareness (\SIrange{57}{78}{\percent} collision). MA-MPC's closed-loop variant cuts collisions to \SIrange{10}{20}{\percent} but the partner-blind reference and the lack of multi-step reverse-in cap SR below \SI{60}{\percent}. HOPE's single-agent DLP training plus open-loop RS terminal approach collapses on DSC3D transfer (\SI{8.4}{\percent} / \SI{6.5}{\percent} SR). GigaFlow and CaRL do not discover multi-step reverse-in within budget, leaving terminal error stuck at \SI{0.5}{\meter}. Diffusion Planner imitates human trajectories but does not internalize the closed-loop constraints arising when its own actions perturb partners (\SI{10.5}{\percent} / \SI{5.8}{\percent} collision). \method's failures concentrate in overcrowded aisles requiring centimeter-level accuracy. Per-source collision breakdowns (vehicle--vehicle vs.\ vehicle--static) for all methods are reported in Appendix~\ref{app:coll-breakdown}.

\begin{figure}[!t]
\centering
\vspace{-8pt}
\includegraphics[width=0.92\linewidth]{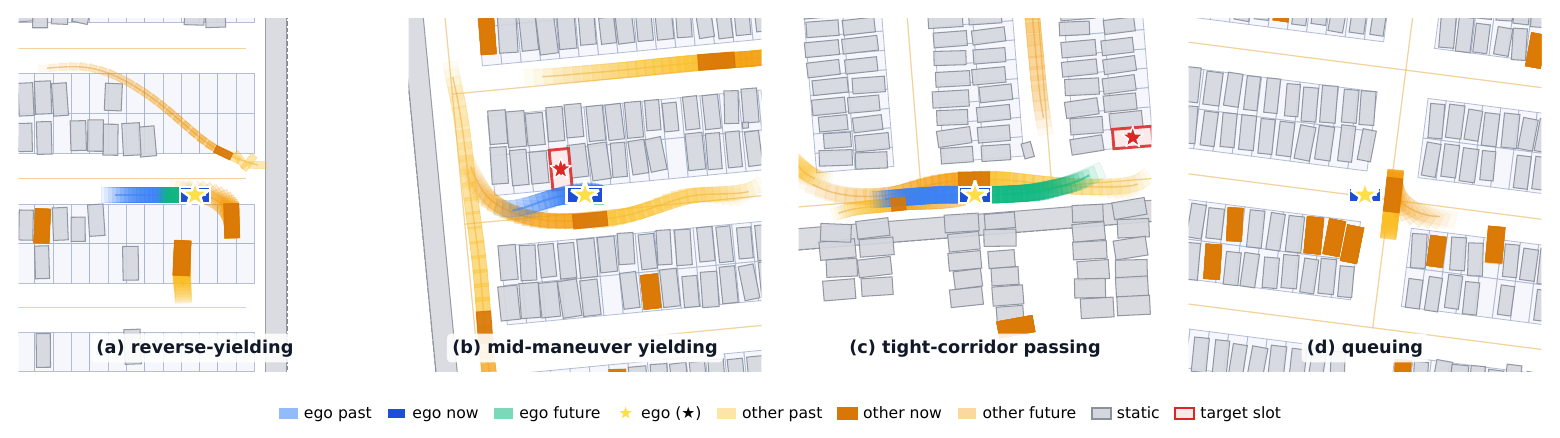}
\vspace{-8pt}
\caption{\small Emergent interaction behaviors of \method on held-out datasets. (a) Reverse-yielding: the ego backs out of an aisle to let an outgoing partner pass. (b) Mid-maneuver yielding: the ego pauses partway into its slot entry to give space to a crossing partner. (c) Tight-corridor passing: two agents traverse a narrow aisle within centimeter-scale clearance. (d) Queuing: agents form an orderly line behind a slow predecessor. The full token legend is shown at the bottom of the figure.}
\label{fig:qualitative}
\vspace{-14pt}
\end{figure}

\textbf{Emergent Multi-agent Behaviors}: \label{sec:emergent} The four behaviors in Fig.~\ref{fig:qualitative} arise under self-play without behavior-specific rewards, and split by the phase in which interaction occurs. The first pair shows that the policy stays responsive to neighbors during the parking maneuver itself, not only during routing, which is the capability that distinguishes \method from prior parking pipelines. In reverse-yielding (a), the ego in its slot-entry maneuver detects an outgoing partner and yields longitudinally by backing out, while the lateral channel stays anchored to the slot frame, jointly enabled by the asymmetric release and self-play training. Mid-maneuver yielding (b) relies on the longitudinal-only release: the residual pauses forward progress when a crossing partner enters its envelope, without abandoning the lateral reference that keeps slot alignment intact. The second pair shows the same responsiveness during normal navigation: in tight-corridor passing (c), self-play has driven the policy to enough precision for two agents to clear each other within centimeter-scale aisle clearance, while in queuing (d), self-play training has produced a yielding behavior behind a slow predecessor without any explicit coordination signal.\looseness=-1

\subsection{Ablation Study}
\label{sec:ablations}
We present a cumulative components analysis and a channel-wise prior-release sensitivity study below. Seven further ablations and analyses (scene-density curriculum, success-criterion sensitivity, self-play partner evaluation, partner-observation noise, path/maneuver distribution, per-source collision breakdown, and inference-time reward-mode control) are in Appendix~\ref{app:ablations}.

\setlength{\intextsep}{2pt}
\begin{wraptable}{r}{0.50\linewidth}
\centering
\scriptsize
\setlength{\tabcolsep}{3pt}
\setlength{\abovecaptionskip}{4pt}
\renewcommand{\arraystretch}{0.9}
\begin{tabular}{l ccc}
\toprule
\textbf{Variant} & \textbf{SR\,[\%]} & \textbf{Coll\,[\%]} & \textbf{PErr\,[m]} \\
\midrule
Vanilla GigaFlow~\cite{cusumano2025robust} & $50.8_{\pm 1.7}$ & $3.8_{\pm 0.7}$  & $0.5$ \\
\;+ Prior                                  & $68.7_{\pm 1.7}$ & $13.3_{\pm 1.2}$ & $0.2$ \\
\;+ Asym.~reliance                         & $78.3_{\pm 1.5}$ & $3.9_{\pm 0.7}$  & $0.5$ \\
\;+ Refinement (full \method)              & $\mathbf{84.7}_{\pm 1.4}$ & $4.1_{\pm 0.7}$ & $\mathbf{0.4}$ \\
\bottomrule
\end{tabular}
\caption{\small Cumulative components analysis on DLP-Reactive ($\pm$\,std over $5$ seeds). Starting from the GigaFlow self-play baseline (DLP-Reactive row of Table~\ref{tab:main-results}), each row adds one \method module: offline kinematic prior, asymmetric channel-wise reliance, and closed-loop refinement.}
\label{tab:enabler-ablation}
\end{wraptable}
\textbf{Cumulative Components}: Starting from the GigaFlow self-play baseline (Tab.~\ref{tab:enabler-ablation}, DLP-Reactive row of Tab.~\ref{tab:main-results}), each row adds one \method module. The offline kinematic prior supplies a geometric reference that sharpens terminal precision and lifts SR by roughly 18 points, but the residual head now tracks a partner-blind plan, and collisions spike from $3.8\%$ to $13.3\%$. Asymmetric reliance then releases the longitudinal channel while anchoring the lateral one, returning collisions to baseline ($3.9\%$) and lifting SR another $\sim$ten points. Precision relaxes here as yielding pulls the policy off the prior, the residual error refinement then absorbs. Closed-loop refinement recovers precision and pushes SR to its $84.7\%$ peak, with a small uptick in collisions from grazing neighbors mid-correction.

\setlength{\intextsep}{12pt plus 2pt minus 2pt}

\begin{table}[!t]
\centering
\small
\setlength{\tabcolsep}{6pt}
\renewcommand{\arraystretch}{1.0}
\begin{tabular}{l cccc}
\toprule
$(\kappa^{\text{acc}},\,\kappa^{\text{steer}})$ & \textbf{SR\,[\%]} & \textbf{Coll\,[\%]} & \textbf{V--V\,[\%]} & \textbf{V--Stat\,[\%]} \\
\midrule
\multicolumn{5}{l}{\textit{Symmetric} ($\kappa^{\text{acc}} = \kappa^{\text{steer}}$)} \\
$(0.40,\,0.40)$ & $70.5_{\pm 1.7}$ & $11.8_{\pm 1.2}$ & $8.4_{\pm 1.0}$ & $3.4_{\pm 0.6}$ \\
$(0.70,\,0.70)$ & $76.3_{\pm 1.5}$ & $8.9_{\pm 1.0}$  & $5.2_{\pm 0.8}$ & $3.7_{\pm 0.6}$ \\
$(0.95,\,0.95)$ & $78.2_{\pm 1.6}$ & $7.6_{\pm 1.0}$  & $2.1_{\pm 0.5}$ & $5.5_{\pm 0.8}$ \\
\midrule
\multicolumn{5}{l}{\textit{Reverse asymmetric} ($\kappa^{\text{steer}} > \kappa^{\text{acc}}$)} \\
$(0.40,\,0.70)$ & $69.3_{\pm 1.7}$ & $11.4_{\pm 1.2}$ & $7.5_{\pm 0.9}$ & $3.9_{\pm 0.6}$ \\
$(0.40,\,0.95)$ & $66.8_{\pm 1.8}$ & $14.4_{\pm 1.4}$ & $8.6_{\pm 1.1}$ & $5.8_{\pm 0.8}$ \\
$(0.70,\,0.95)$ & $73.5_{\pm 1.6}$ & $9.7_{\pm 1.1}$  & $4.6_{\pm 0.7}$ & $5.1_{\pm 0.7}$ \\
\midrule
\multicolumn{5}{l}{\textit{Correct asymmetric} ($\kappa^{\text{acc}} > \kappa^{\text{steer}}$)} \\
$(0.70,\,0.40)$ & $80.5_{\pm 1.5}$ & $6.5_{\pm 0.9}$  & $4.2_{\pm 0.7}$ & $2.3_{\pm 0.5}$ \\
$(0.95,\,0.70)$ & $81.4_{\pm 1.5}$ & $6.4_{\pm 0.9}$  & $2.6_{\pm 0.6}$ & $3.8_{\pm 0.6}$ \\
$\mathbf{(0.95,\,0.40)}$ \textbf{(default)} & $\mathbf{84.7}_{\pm 1.4}$ & $\mathbf{4.1}_{\pm 0.7}$ & $\mathbf{2.4}_{\pm 0.5}$ & $\mathbf{1.7}_{\pm 0.4}$ \\
\bottomrule
\end{tabular}
\caption{\small Two-dimensional sweep of the channel-wise prior-release rates $(\kappa^{\text{acc}}, \kappa^{\text{steer}})$ on DLP-Reactive ($\pm$\,std over $5$ seeds), evaluated at $\kappa \in \{0.40, 0.70, 0.95\}$ per channel. Rows are grouped into symmetric, reverse-asymmetric ($\kappa^{\text{steer}} > \kappa^{\text{acc}}$), and correct-asymmetric ($\kappa^{\text{acc}} > \kappa^{\text{steer}}$) blocks. Collisions are split into vehicle--vehicle (V--V) and vehicle--static (V--Stat) contacts. Bold marks the configuration used throughout the rest of the paper.}
\label{tab:asym-ablation}
\end{table}

\textbf{Channel-wise Prior-Release Sensitivity}: Table~\ref{tab:asym-ablation} sweeps the release rates $(\kappa^{\text{acc}}, \kappa^{\text{steer}})$ of Eq.~\ref{eq:split-reliance} over $\{0.40, 0.70, 0.95\}$ per channel on DLP-Reactive, splitting collisions into vehicle--vehicle (V--V) and vehicle--static (V--Stat) contacts. The sweep separates release magnitude from release direction. Along the symmetric diagonal, a faster release trades V--V contacts ($8.4\%$ to $2.1\%$, more longitudinal yielding) for V--Stat contacts ($3.4\%$ to $5.5\%$, more lateral drift off the slot reference), and SR plateaus in the $70$--$78\%$ band. The reverse-asymmetric block degrades both channels, since the longitudinal anchor blocks yielding while the unanchored lateral channel drifts, and SR collapses to $66.8\%$ at $(0.40, 0.95)$. The correct-asymmetric block breaks the trade-off: every entry beats every symmetric and reverse-asymmetric configuration on SR, and the default $(0.95, 0.40)$ nearly matches the best V--V rate of the diagonal ($2.4\%$ vs.\ $2.1\%$) while reaching the lowest V--Stat rate ($1.7\%$) and the highest SR ($84.7\%$). The direction of the asymmetry, not the release magnitude, is therefore the operative design choice.
\section{Conclusion}
\label{sec:conclusion}

We introduce reactive parking, the closed-loop multi-agent extension of automated parking, and propose \method, a self-play RL approach combining an offline kinematic prior, a residual policy with reactive yielding under partner threat, and a closed-loop refinement layer, all trained within a single parameter set under multi-agent self-play. \method substantially outperforms classical, imitation, and pure large-scale RL baselines on success rate under both partner reaction modes, with collision rates within a percentage point of the best large-scale RL baseline, and exhibits emergent interaction behaviors without behavior-specific rewards. Beyond the immediate result, \method instantiates a general principle: a geometric prior should be released to match a task's anisotropic tolerance, anchored where tolerance is tight and relaxed where it is loose. We expect this principle to transfer to other multi-agent settings where strong geometric priors coexist with partner dynamics that cannot be pre-planned, provided the task's precision tolerance aligns with distinct action channels.
\section{Limitations}
\label{sec:limitations}

Compute constraints prevented scaling along orthogonal axes such as larger lot populations (the densest training tier was capped at \num{32} controlled agents per scene), longer training horizons, and broader hyperparameter sweeps. \method relies on an offline kinematic prior to attain terminal precision, and we have not yet explored a more end-to-end alternative that reaches the same accuracy without this external scaffold. Terminal precision is additionally bounded by the $91$-way discrete action grid imposed by the PufferDrive simulator interface. A continuous-action substrate would further tighten PErr. Evaluation is sim-only within this submission's budget. Real-world deployment would further require a passive safety layer given the tight accuracy tolerances of production-grade parking. \looseness=-1


\bibliography{reference}

\clearpage
\appendix
\begin{center}
{\Large\bfseries CoPark: Learning Reactive Parking via Self-Play\\[0.4em]Appendix\par}
\end{center}
\vspace{1.5em}
\section{Training Details}
\label{app:training}

This section collects the implementation details supporting the training setup of Sec.~\ref{sec:setup}. We describe the parking-lot layouts and scene sampling pipeline (Appendix~\ref{app:scene-prep}), the observation interface and policy architecture (Appendix~\ref{app:obs-architecture}), the reward composition with per-agent conditioning (Appendix~\ref{app:reward}), and the optimization recipe (Appendix~\ref{app:hyperparameters}). A visualization of the observation features and the resulting policy distribution on a representative episode closes the section (Appendix~\ref{app:obs-action-viz}).

\subsection{Training Maps and Scene Construction}
\label{app:scene-prep}

Training uses six custom parking-lot layouts, all normalized into a common clean-map representation that declares the drivable region, slot poses with footprints, a directed lane graph, and static-obstacle polygons. DLP~\cite{DLP} and DSC3D~\cite{DSC3D} are converted into the same representation for zero-shot evaluation (Appendix~\ref{app:datasets}). The six layouts span a range of parking-lot footprints and aisle topologies, providing the layout diversity that \method's policy generalizes across before zero-shot transfer.

Each scene samples a static-occupancy fraction and a controlled-agent count from a three-tier stratification with occupancies $0.25$, $0.50$, and $0.75$, up to $32$ controlled vehicles per scene, and the densest tier weighted at $50\%$. Static parked vehicles are placed uniformly at random in unoccupied slots and merged into the scene's \texttt{road\_edge} geometry, a workaround required by the maximum-controlled-agent cap of the PufferDrive simulator. A source tag is retained on every merged static vehicle so that vehicle--vehicle, vehicle--static, and off-road terminal events remain distinguishable in downstream analysis (Appendix~\ref{app:coll-breakdown}). For each controlled agent, an initial pose and an injective slot assignment are drawn offline under a kinematic-reachability check and held fixed throughout the episode.

\subsection{Observation Space and Policy Architecture}
\label{app:obs-architecture}

The policy is a single parameter-sharing categorical actor $\pi_\theta(a \mid s)$ over the $91$-way discrete action grid ($7$ acceleration levels by $13$ steering angles), with a shared backbone fusing five feature blocks. The ego block carries the agent's kinematic state. The partner block carries the relative state of the $K=8$ nearest neighbors as oriented bounding boxes in ego frame; $K=8$ covers the $99$th percentile of nearby-vehicle counts within the observation range across training scenes, and larger values of $K$ provided no measurable improvement in pilot studies. The road block carries polyline tokens for lane centerlines and parking-lot boundaries, sampled around the ego in ego frame under per-class token budgets. The reward-conditioning block carries the per-agent weights $(w_g, w_c)$ defined in Appendix~\ref{app:reward}. The tail block carries the $10$-dimensional normalized scalar summary of the closing-threat and static-clearance signals, the signed slot-frame errors, the Stanley reference command, the teacher cross-track and heading errors, and the ego signed velocity.

The Stanley reference enters the actor as a channel-wise log-prior added to the policy logits (Eq.~\ref{eq:policy}), with the reliance scalars contracting along the saturated threat signal $\rho(s)$ (Eq.~\ref{eq:split-reliance}). The residual logit head $z_\theta(s, a)$ is bounded so that the prior pull retains effect when $\rho(s) \to 0$. The Stanley controller writes nothing to the simulator; its action enters the policy only through the log-prior channel. The slot-frame errors and the Stanley command are zeroed during the navigation phase and become informative only after the agent transitions to the parking-maneuver phase, so that the phase-conditional prior of Eq.~\ref{eq:policy} fires only when geometrically meaningful.

\subsection{Reward Composition and Per-Agent Conditioning}
\label{app:reward}

The per-agent reward decomposes into four groups (Table~\ref{tab:reward-terms}). A sparse terminal group fires only on episode-ending events. A phase-independent dense safety group is active throughout the episode. Two phase-dependent dense shaping groups take effect during navigation and during the parking maneuver respectively. Contact is terminal and removes the agent from the rollout, so no in-contact dense penalty is needed.

\begin{table}[!htbp]
\centering
\small
\setlength{\tabcolsep}{6pt}
\renewcommand{\arraystretch}{1.10}
\begin{tabular}{l l l}
\toprule
\textbf{Group} & \textbf{Term} & \textbf{When active} \\
\midrule
\multirow{3}{*}{Sparse terminal}
 & Success bonus                                              & on success (Eq.~\ref{eq:success}) \\
 & Terminal collision penalty                                 & on V--V or V--Stat contact \\
 & Navigation$\to$Maneuver handoff bonus                      & on entering maneuver phase \\
\midrule
\multirow{2}{*}{\makecell[l]{Dense safety\\(both phases)}}
 & Closing-threat penalty $-\beta_{\text{dyn}}\cdot\text{threat}_i$ & continuous in $\text{threat}_i$ (Eq.~\ref{eq:threat}) \\
 & Static-clearance penalty                                   & if ego $<$ clearance margin to static edge \\
\midrule
\multirow{3}{*}{Navigation shaping}
 & Lane-graph progress                                        & per step along $\mathcal{G}$ \\
 & Subgoal bonus                                              & on reaching each $g_{i,k}$ \\
 & Triangular speed-target reward                             & per step under target speed band \\
\midrule
\multirow{3}{*}{Maneuver shaping}
 & Arc-along-plan progress                                    & per step along $\tau_i$ \\
 & Threat-modulated imitation $r^{\text{imit}}_{i,t}$ (Eq.~\ref{eq:imitation}) & scaled by $(1-\rho(s))\,w_{\text{trust}}(s)$ \\
 & Refinement reward $r^{\text{ref}}_{i,t}$ (Eq.~\ref{eq:refine-reward}) & during closed-loop refinement \\
\bottomrule
\end{tabular}
\caption{\small Per-agent reward decomposition. Numerical values of the weights ($\beta_{\text{dyn}}, \alpha_{\text{imit}}, \beta_{\text{red}}, \beta_{\text{hold}}$) and of the prior-modulation hyperparameters (Eqs.~\ref{eq:trust}, \ref{eq:split-reliance}) are listed in Table~\ref{tab:hparams}.}
\label{tab:reward-terms}
\end{table}

The numerical values of the method-side hyperparameters introduced in Sec.~\ref{sec:method} are collected in Table~\ref{tab:hparams}. The static-clearance gate slope $\beta_{\text{stat}}$ is set steep so that $\sigma(\beta_{\text{stat}}(d_{\text{stat}}-d_0))$ approximates a hard step at $d_0$.

\begin{table}[!htbp]
\centering
\small
\setlength{\tabcolsep}{8pt}
\renewcommand{\arraystretch}{1.05}
\begin{tabular}{l l l}
\toprule
\textbf{Symbol} & \textbf{Description} & \textbf{Value} \\
\midrule
\multicolumn{3}{l}{Threat signal and prior modulation (Eqs.~\ref{eq:threat}, \ref{eq:split-reliance})} \\
$d_{\text{decay}}$                          & threat range decay constant                 & $4.5$\,m \\
$\tau_\rho$                                 & threat squash scale ($\rho$ saturation)     & $1.0$\,m/s \\
$\alpha_{\min},\,\alpha_{\max}$             & reliance scalar range                       & $1.0,\;4.0$ \\
$\kappa^{\text{acc}},\,\kappa^{\text{steer}}$ & channel-wise release rates                & $0.95,\;0.40$ \\
\midrule
\multicolumn{3}{l}{Trust-weight components (Eq.~\ref{eq:trust})} \\
$\sigma_{\text{ct}}$                        & cross-track suppression scale               & $1.5$\,m \\
$\beta_{\text{stat}}$                       & static-clearance gate slope                 & $20$\,m$^{-1}$ \\
$d_0$                                       & static-clearance threshold                  & $0.35$\,m \\
\midrule
\multicolumn{3}{l}{Reward weights (Eqs.~\ref{eq:imitation}, \ref{eq:refine-reward})} \\
$\alpha_{\text{imit}}$                      & threat-modulated imitation weight           & $0.20$ \\
$\sigma_{\text{imit}}$                      & imitation command-match scale               & $0.50$ \\
$\beta_{\text{dyn}}$                        & dynamic threat penalty weight               & $0.10$ \\
$\beta_{\text{red}}$                        & refinement directed-gradient gain          & $0.25$ \\
$\beta_{\text{hold}}$                       & refinement hold bonus                       & $0.08$ \\
$\eta$                                      & refinement tanh saturation scale            & $0.10$\,m \\
\bottomrule
\end{tabular}
\caption{\small Numerical values of the method-side hyperparameters of Sec.~\ref{sec:method}.}
\label{tab:hparams}
\end{table}

Each agent additionally receives a pair of conditioning weights $(w_{\text{g}}, w_{\text{c}})$ that scale the goal-related and collision-related reward components respectively, following the recipe of GigaFlow~\cite{cusumano2025robust}. The weights are sampled per agent at episode start from $w_{\text{g}} \sim \mathcal{U}[0.5, 2.0]$ and $w_{\text{c}} \sim \mathcal{U}[0.5, 3.0]$, so that a single policy spans a continuum of operational styles from cautious-and-precise to aggressive-and-fast within the same parameter set. The same range is later swept at inference time in Appendix~\ref{app:reward-mode} to demonstrate operating-point control on a single trained checkpoint.

\subsection{Optimization Recipe and Training Hyperparameters}
\label{app:hyperparameters}

Training uses the PPO~\cite{schulman2017proximal} variant in PufferLib~\cite{suarez2024pufferlib} with V-trace~\cite{espeholt2018impala} advantages under parameter sharing, for $4 \times 10^9$ environment steps on $2$ NVIDIA RTX~A6000 GPUs, sustaining approximately $120$k simulator steps per second at episode length $400$. The three-tier scene-density curriculum of Appendix~\ref{app:scene-prep} schedules occupancies $0.25$, $0.50$, and $0.75$ with the densest tier sampled at $50\%$, exposing the policy to progressively denser interactions over the course of training. The same parameter set $\theta$ is shared across both phases, all agents, and all episodes; phase information enters only through the phase indicator $\phi_{i,t}$ in the observation and through the phase-conditional prior of Eq.~\ref{eq:policy}. Completed agents are removed from the rollout on termination, so the active-agent count varies during an episode.

\subsection{Observation and Policy-Distribution Visualization}
\label{app:obs-action-viz}

\begin{figure}[!htbp]
\centering
\includegraphics[width=0.93\linewidth]{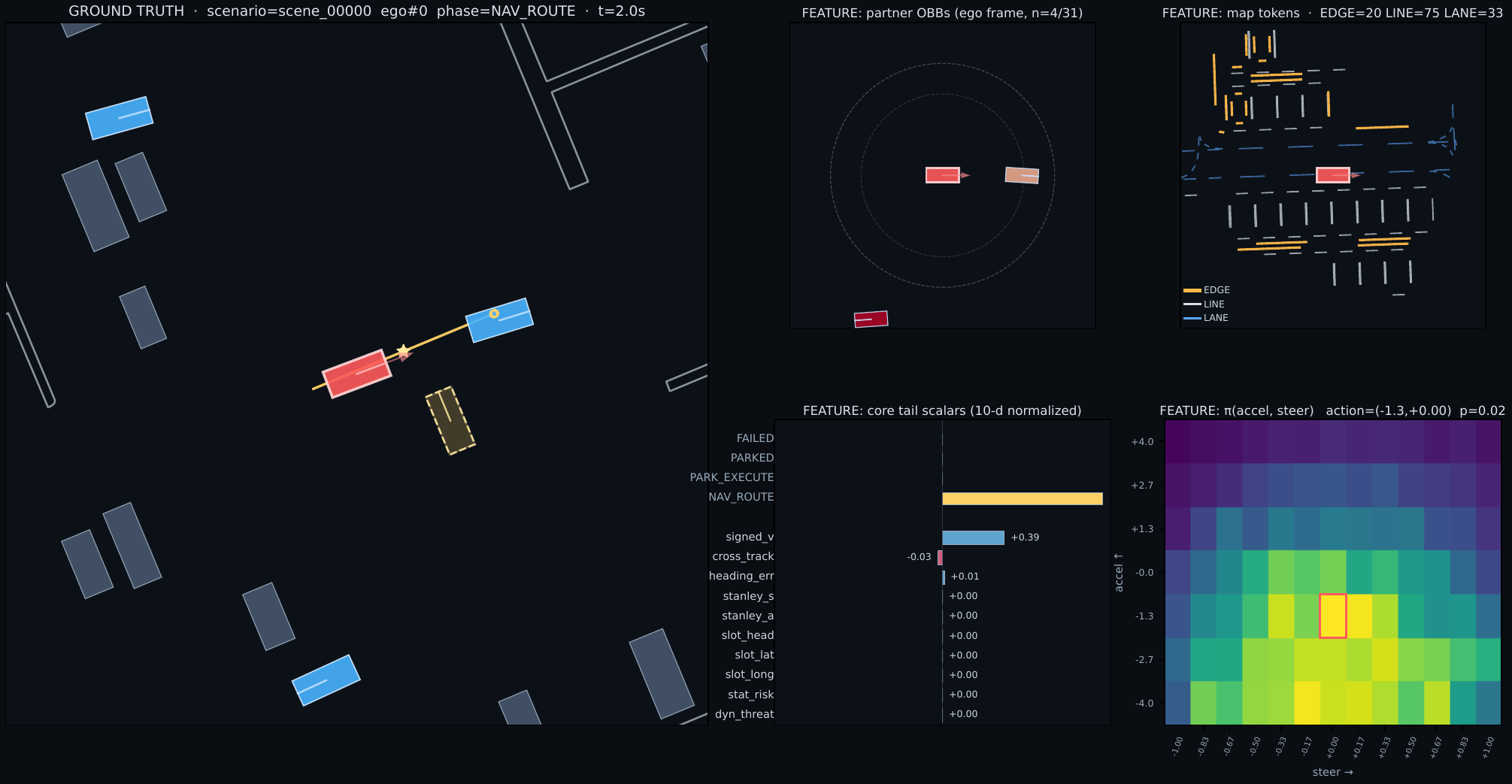}\\[2pt]
\includegraphics[width=0.93\linewidth]{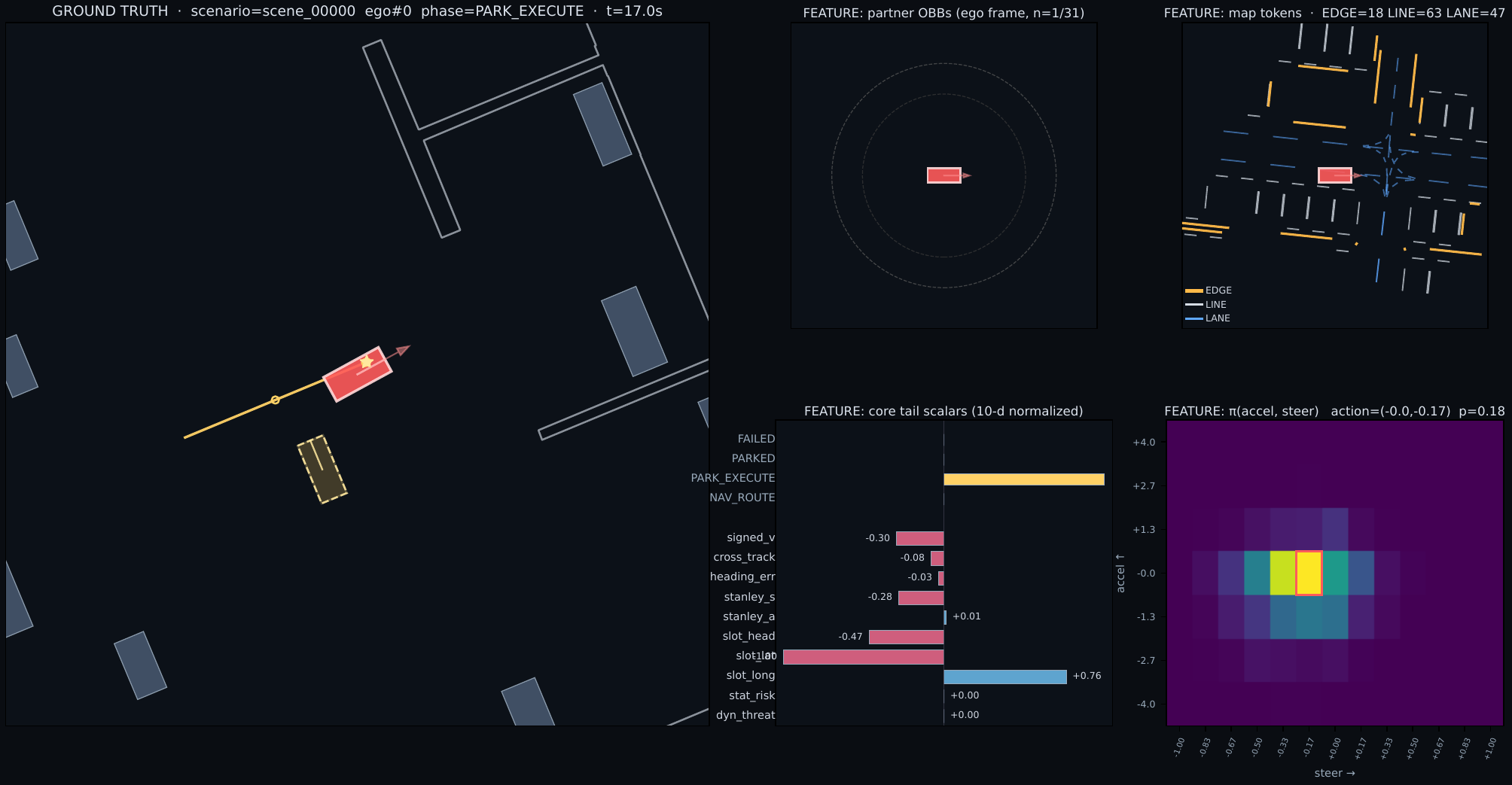}\\[2pt]
\includegraphics[width=0.93\linewidth]{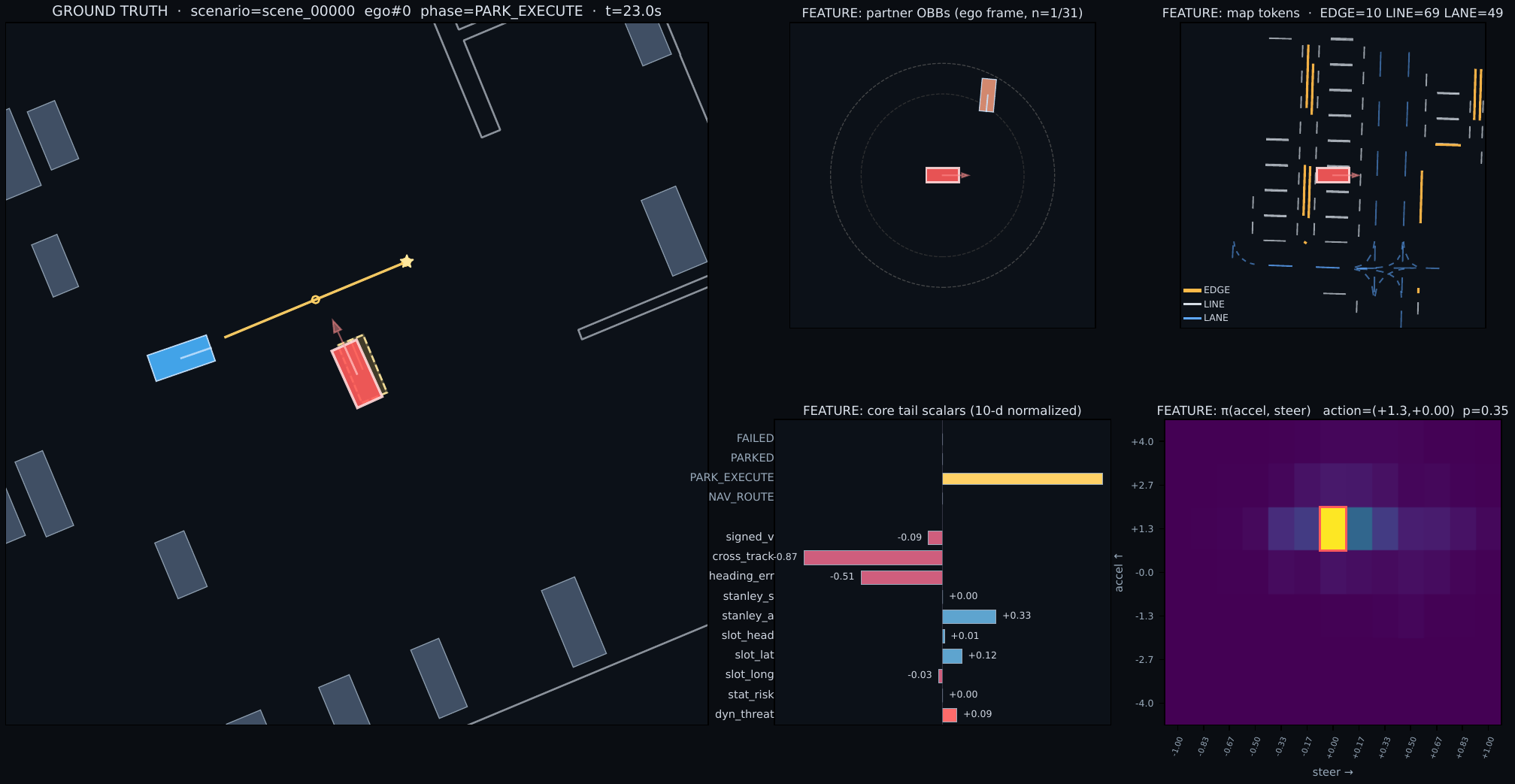}
\caption{\small Policy observation and action distribution at three phases of a representative training episode on a custom parking-lot layout (same agent, same scene). The top row at $t=2.0$\,s shows phase \texttt{NAV\_ROUTE}, free-roaming navigation with $4/31$ partners in range. The middle row at $t=17.0$\,s shows phase \texttt{PARK\_EXECUTE} at maneuver onset near the preparation pose, with large slot-frame offsets ($\text{slot\_long}=+0.76$, $\text{slot\_lat}=-1.00$). The bottom row at $t=23.0$\,s shows phase \texttt{PARK\_EXECUTE} late maneuver, with the ego nearly aligned to its slot ($\text{slot\_long}=-0.03$, $\text{slot\_lat}=+0.12$). Each row uses the same five-panel layout described in the text.}
\label{fig:feature-viz}
\end{figure}

To make concrete what \method's policy reads and writes, Fig.~\ref{fig:feature-viz} visualizes a single representative training episode on one of the custom parking-lot layouts at three phase-defining instants: free-roaming navigation, the start of the parking maneuver near the preparation pose, and the late maneuver as the ego closes onto its slot. Every row shares the same five-panel layout, described as follows.

\begin{itemize}[topsep=2pt,itemsep=1pt,leftmargin=*]
    \item \textbf{Ground truth (left).} Bird’s-eye view of the lot. The ego is red, static parked vehicles slate, other controlled agents blue, the assigned target slot a yellow dashed rectangle, the navigation-route endpoint (preparation pose $p_i^{\text{prep}}$) a yellow star, and the Hybrid~A$^*$ plan with its Stanley reference appear as the yellow line with circle waypoint markers.
    \item \textbf{Partner OBBs (top-mid).} The partner oriented bounding boxes the policy actually sees in ego frame, after $K$-nearest-neighbors cropping ($K=8$). The annotation $n/N$ at the panel header reports the $n$ partners retained within the observation range out of the $N$ total present in the scene, and the dashed rings indicate the observation range.
    \item \textbf{Map tokens (top-right).} The polyline tokens (lane centerlines and parking-lot boundaries) sampled around the ego in ego frame, with the per-class token budget actually used (LANE/BOUND counts).
    \item \textbf{Core tail scalars (bottom-mid).} The $10$-dimensional normalized scalar tail of the observation: the closing-threat and static-clearance scalars, the signed slot-frame errors ($\text{slot\_long}, \text{slot\_lat}, \text{slot\_head}$), the Stanley target ($\text{stanley\_a}, \text{stanley\_s}$), the teacher cross-track and heading errors, and the ego signed velocity.
    \item \textbf{Policy distribution $\pi_\theta(a\mid s)$ (bottom-right).} The full categorical distribution over the $7{\times}13$ acceleration--steering action grid, with the sampled action outlined in red and its probability $p$ printed at the panel header.
\end{itemize}

Three structural properties of the observation are immediately visible from the comparison. (i) The phase one-hot switches cleanly from \texttt{NAV\_ROUTE} (top) to \texttt{PARK\_EXECUTE} (middle, bottom). The slot-frame errors and the Stanley target are exactly zero during navigation and become informative only once the parking-maneuver phase starts, exactly as required by the phase-conditional prior of Eq.~\ref{eq:policy} and the threat-modulated imitation reward of Eq.~\ref{eq:imitation}. (ii) Partner density falls from $4/31$ in the open navigation phase to $1/31$ inside the dense aisle while map-token counts stay roughly constant (LANE $33$--$49$, BOUND $10$--$20$), confirming that the cropping budget is exercised by partners rather than map geometry. (iii) The policy distribution evolves with the phase. It is broad over the action grid during navigation (sampled-action probability $p=0.02$, reflecting genuine in-traffic uncertainty), tightens to a localized mode at maneuver onset ($p=0.18$), and sharpens further into a near-deterministic forward step at late maneuver ($p=0.35$). The progression shows how the residual head retains exploratory entropy while the prior is dormant and concentrates onto the prior-anchored geometric reference as the slot frame becomes informative.

\section{Evaluation Protocol Details}
\label{app:benchmark}

This section provides the implementation details that ensure all methods are evaluated under identical conditions. We describe the source datasets and episode construction (Appendix~\ref{app:datasets}), the success criterion and tolerances (Appendix~\ref{app:termination}), the three partner modes used in evaluation (Appendix~\ref{app:partner-modes}), the evaluation harness and shared action interface (Appendix~\ref{app:planner-protocol}), and the adaptation procedure for each baseline (Appendix~\ref{app:baselines}).

\subsection{Datasets and Episode Construction}
\label{app:datasets}

We integrate the Dragon Lake Parking dataset (DLP)~\cite{DLP} and the DeepScenario Open 3D dataset (DSC3D)~\cite{DSC3D} into the PufferDrive~\cite{pufferdrive2025github} simulator under the formulation of Sec.~\ref{sec:problem-formulation}. Both datasets are converted into the same clean-map representation used at training (Appendix~\ref{app:scene-prep}), comprising a drivable region, slot poses with footprints, a directed lane graph, and static-obstacle polygons. Every source-data parking trajectory yields one ego-episode in which the ego is the original parking driver and the assigned slot is its final logged pose. Non-ego agents present in the source data are converted into scripted partners (Appendix~\ref{app:partner-modes}). The two datasets differ in scene structure. DSC3D mixes through-road segments with parking-garage areas, whereas \method's training layouts are pure parking-garage environments, so the lane-graph topology DSC3D presents at evaluation departs from the structure seen during self-play. This structural gap accounts for the lower zero-shot SR on DSC3D than on DLP in Table~\ref{tab:main-results}.

\subsection{Success Criterion and Tolerances}
\label{app:termination}

An episode ends in success, collision, off-road, refinement timeout (when the budget $K_{\text{ref}}^{\max}$ is exhausted), or horizon timeout. Success requires Eq.~\ref{eq:success} to hold for $K_s$ consecutive steps in the slot-aligned frame. Numerical tolerances and horizons are listed in Table~\ref{tab:eval-params}. The position criterion is the Euclidean slot-frame distance $\|e_{i,t}^{\text{slot}}\|_2 \le \Delta_d$, so the reported PErr in Table~\ref{tab:main-results} and the success gate share a single geometric notion. The $10^{\circ}$ heading tolerance with modulo-$\pi$ wrapping accepts both forward-in and reverse-in slot entries.

\begin{table}[!htbp]
\centering
\small
\setlength{\tabcolsep}{10pt}
\renewcommand{\arraystretch}{1.1}
\begin{tabular}{ll}
\toprule
\textbf{Parameter} & \textbf{Value} \\
\midrule
Simulator step size & $0.1$\,s ($10$\,Hz) \\
Training / evaluation horizon & $400$ / $1{,}800$ steps \\
Refinement budget $K_{\text{ref}}^{\max}$ & $80$ steps \\
Success tolerance, position $\Delta_d$ & $0.8$\,m (Euclidean distance to slot center) \\
Success tolerance, heading $\Delta_\psi$ & $0.1745$\,rad $\approx 10^{\circ}$ (modulo $\pi$) \\
Success tolerance, speed $\Delta_v$ & $0.35$\,m/s \\
Required hold duration $K_s$ & $5$ consecutive steps \\
Controlled-agent dimensions (L $\times$ W, wheelbase) & $3.2 \times 1.4$\,m, $2.1$\,m \\
\bottomrule
\end{tabular}
\caption{\small Parameters of the reactive-parking evaluation protocol in PufferDrive.}
\label{tab:eval-params}
\end{table}

\subsection{Partner Modes}
\label{app:partner-modes}

Following the closed-loop protocol of GigaFlow~\cite{cusumano2025robust}, each method controls a single ego in every episode while all non-ego partners advance under one of three scripted modes. The reactive mode advances each non-ego partner along its logged route using IDM~\cite{treiber2000idm}, with the longitudinal command decelerating in response to the ego's proximity. The non-reactive mode replays each non-ego partner's logged trajectory regardless of the ego's behavior, producing the harder collision-avoidance condition because partners do not yield. The self-play mode advances every non-ego partner under the same \method policy as the ego, with each partner pursuing its own assigned slot. The reactive and non-reactive modes form the headline two partner-mode blocks of Table~\ref{tab:main-results}. The self-play mode is reported separately in Appendix~\ref{app:self-play-eval}.

\subsection{Evaluation Harness and Action Interface}
\label{app:planner-protocol}

At every $0.1$\,s step the harness collects the joint state, constructs the ego observation, and queries the method under test for one of the $91$ discrete acceleration--steering primitives. The $7 \times 13$ grid is mandated by the PufferDrive simulator's action interface and sets an inherent floor on terminal precision across all methods; \method's sub-meter PErr in Table~\ref{tab:main-results} operates near this discretization limit. All baselines that natively produce continuous commands are projected onto the same grid by nearest-grid-cell assignment at each step. The Stanley-tracked commands from classical planners, the MPC commands of MA-MPC, HOPE's policy outputs, and Diffusion Planner's sampled actions are all discretized this way before being applied to the simulator. GigaFlow and CaRL already operate over the same discrete grid in our reimplementation. The shared action interface and shared contact model ensure that performance differences in Table~\ref{tab:main-results} reflect policy quality rather than action-space mismatch.

\subsection{Baseline Adaptations}
\label{app:baselines}

\textbf{Classical planners}: RS~\cite{Reeds1990OPTIMALPF} re-plans a Reeds--Shepp curve from the agent's current pose to its assigned slot at every step. Hybrid~A$^{*}$~\cite{Dolgov2010PathPF} uses the same plan library that \method uses offline, but invoked online and tracked open-loop by Stanley control without the learned residual.

\textbf{MA-MPC}: Each ego runs a receding-horizon predictive controller over the same Hybrid~A$^{*}$ reference path used in the Hybrid~A$^{*}$+Stanley row, with dynamic-obstacle constraints constructed from observed partner positions and a constant-velocity prediction over the horizon. The continuous MPC command is projected onto the $91$-way action grid (Appendix~\ref{app:planner-protocol}) before being applied. The framework follows the multi-vehicle MPC recipe of~\cite{9091894}, adapted from V2I-coordinated to decentralized per-agent execution.

\textbf{HOPE}: HOPE~\cite{jiang2025hope} couples a learned RL policy with rule-based Reeds--Shepp curves via action masking. We reuse the released checkpoint adapted to the PufferDrive observation format. Since HOPE is trained on parking scenes derived from DLP layouts, its evaluation on DLP is not strictly zero-shot, while on DSC3D it is.

\textbf{GigaFlow and CaRL}: GigaFlow~\cite{cusumano2025robust} and CaRL~\cite{jaeger2025carl} are two large-scale RL recipes trained on the same six maps and same compute budget as \method, but without the \method components. The two differ in reward complexity. GigaFlow uses a richer per-agent reward continuum, while CaRL uses a deliberately simple reward. Both plateau on success rate well below \method in the dense low-speed parking domain.

\textbf{Diffusion Planner}: Diffusion Planner~\cite{zheng2025diffusion} is trained in its standard imitation regime directly on DLP and DSC3D human demonstrations using fixed-time-budget trajectory windows, with the original sampler, guidance, and projection pipeline retained. This baseline sees the real evaluation datasets at training time, while \method does not. \method nonetheless outperforms it across all four (dataset, partner-mode) configurations.

\section{Ablations and Extended Analyses}
\label{app:ablations}

This section complements the cumulative components analysis of Sec.~\ref{sec:ablations} with deeper studies organized into four groups. A method-design ablation isolates the scene-density curriculum (Appendix~\ref{app:curriculum}). Three robustness analyses probe success-criterion sensitivity (Appendix~\ref{app:precision-sweep}), the self-play partner mode (Appendix~\ref{app:self-play-eval}), and partner-observation noise (Appendix~\ref{app:noise-robustness}). Two behavioral analyses characterize trajectory naturalism against human demonstrations (Appendix~\ref{app:path-manv-dist}) and the per-source collision breakdown across methods (Appendix~\ref{app:coll-breakdown}). A final operating-point analysis demonstrates inference-time control through per-agent reward conditioning (Appendix~\ref{app:reward-mode}).

\subsection{Scene-Density Curriculum Ablation}
\label{app:curriculum}

The scene-density curriculum of Appendix~\ref{app:scene-prep} stratifies training over three occupancy tiers ($0.25$, $0.50$, $0.75$) with the densest tier weighted at $50\%$. Replacing this stratified schedule with a uniform mixed-density distribution from the start, where occupancy is sampled with equal weight across the three tiers throughout training, drops final SR on DLP-Reactive by approximately $14.5$ points (from $84.7\%$ to roughly $70.2\%$). The mechanism is that the policy is overwhelmed by dense partners before it has time to learn navigation, leaving incomplete recovery within the same compute budget. The asymmetric weighting in the stratified schedule is therefore necessary for the policy to acquire navigation competence before being exposed to the densest interaction regime.

\subsection{Success-Criterion Sensitivity}
\label{app:precision-sweep}

The headline tolerance $(\Delta_d, \Delta_\psi) = (0.8\,\text{m}, 10^\circ)$ in Table~\ref{tab:eval-params} reflects practical parking-slot fit-in clearance. Table~\ref{tab:precision-sweep} examines how \method's SR responds to progressively tighter criteria. All other evaluation settings and the trained checkpoint are unchanged; only the success gate is reapplied to the same rollouts.

\begin{table}[!htbp]
\centering
\small
\setlength{\tabcolsep}{8pt}
\renewcommand{\arraystretch}{1.10}
\begin{tabular}{l cc}
\toprule
\textbf{Tolerance} $(\Delta_d, \Delta_\psi)$ & \textbf{DLP-Reactive SR\,[\%]} & \textbf{DSC3D-Reactive SR\,[\%]} \\
\midrule
$(0.8\,\text{m},\,10^\circ)$ (default) & $84.7_{\pm 1.4}$ & $75.8_{\pm 1.7}$ \\
$(0.6\,\text{m},\,8^\circ)$            & $80.5_{\pm 1.5}$ & $71.8_{\pm 1.7}$ \\
$(0.4\,\text{m},\,6^\circ)$            & $62.3_{\pm 1.8}$ & $54.2_{\pm 1.9}$ \\
$(0.2\,\text{m},\,4^\circ)$            & $28.4_{\pm 2.1}$ & $23.6_{\pm 2.2}$ \\
\bottomrule
\end{tabular}
\caption{\small \method's SR under progressively tighter success criteria ($\pm$\,std over $5$ training seeds). The same trained checkpoint is reused; only the post-rollout success gate is varied.}
\label{tab:precision-sweep}
\end{table}

Degradation is gentle from $(0.8, 10^\circ)$ to $(0.6, 8^\circ)$ where the criterion still sits above the mean terminal error of Table~\ref{tab:main-results} (PErr $= 0.4$\,m, HErr $= 3.9^\circ$). The drop steepens at $(0.4, 6^\circ)$ where the criterion bisects the terminal-error distribution, and at $(0.2, 4^\circ)$ where the criterion approaches the discretization limit of the $91$-way action grid on which the refinement layer operates. Classical planners, which achieve decimeter-level PErr (Table~\ref{tab:main-results}), do not benefit from tighter criteria in practice because their absolute SR remains bounded by open-loop collisions.

\subsection{Self-Play Partner Evaluation}
\label{app:self-play-eval}

The reactive and non-reactive partner modes of Appendix~\ref{app:partner-modes} bracket the partner-difficulty spectrum, but neither uses a partner that is itself an active multi-agent planner. We re-evaluate the same trained checkpoint on DLP with all non-ego partners advanced by the same \method policy, so that every vehicle in the lot pursues its own assigned slot under the same residual-prior controller. The result (Table~\ref{tab:self-play-eval}) sits between the two scripted partner modes on SR and below both on collision rate.

\begin{table}[!htbp]
\centering
\small
\setlength{\tabcolsep}{8pt}
\renewcommand{\arraystretch}{1.10}
\begin{tabular}{l cccc}
\toprule
\textbf{Partner mode (DLP)} & \textbf{SR\,[\%]} & \textbf{PErr\,[m]} & \textbf{HErr\,[$^\circ$]} & \textbf{Coll\,[\%]} \\
\midrule
IDM (reactive, default)     & $\mathbf{84.7}_{\pm 1.4}$ & $0.4_{\pm 0.0}$ & $3.9_{\pm 0.2}$ & $4.1_{\pm 0.7}$ \\
Log replay (non-reactive)   & $79.7_{\pm 1.6}$          & $0.4_{\pm 0.0}$ & $4.0_{\pm 0.2}$ & $6.0_{\pm 0.8}$ \\
\textbf{\method (self-play)}& $80.4_{\pm 1.6}$          & $0.4_{\pm 0.0}$ & $4.0_{\pm 0.2}$ & $\mathbf{3.6}_{\pm 0.6}$ \\
\bottomrule
\end{tabular}
\caption{\small Evaluation on DLP with three partner modes ($\pm$\,std over $5$ training seeds). The \method-partner row uses the same trained checkpoint as the ego on every non-ego vehicle.}
\label{tab:self-play-eval}
\end{table}

SR drops $4.3$ points relative to IDM-reactive because \method partners actively pursue their own slots and therefore compete for aisle space that IDM never claims, making genuine conflict cases more frequent. The same mechanism, however, makes them mutually responsive: collisions fall from $4.1\%$ (IDM) to $3.6\%$ thanks to two-sided yielding, lower than either scripted mode. Terminal precision is unchanged because the closed-loop refinement layer operates in the slot-aligned frame, where partner identity does not enter. The result confirms that \method's self-play training transfers to interactions with learned partners, not only to scripted ones, while still leaving headroom above all baselines in Table~\ref{tab:main-results}.

\subsection{Robustness to Partner-Observation Noise}
\label{app:noise-robustness}

The policy is trained on clean simulator observations. To probe sensitivity to the partner-perception inaccuracies expected at deployment, we re-evaluate the same trained checkpoint with isotropic Gaussian noise added to every partner oriented bounding box at inference time, perturbing both position (independently per ego-frame axis) and heading. Three noise levels span a representative range from a high-quality LiDAR detection regime to a stressed sim-to-real condition, and the clean configuration is included as the reference. The training procedure and trained weights are unchanged across rows.

\begin{table}[!htbp]
\centering
\small
\setlength{\tabcolsep}{6pt}
\renewcommand{\arraystretch}{1.0}
\begin{tabular}{l cccc}
\toprule
\textbf{Noise level} ($\sigma_{\text{pos}}$, $\sigma_{\text{head}}$) & \textbf{SR\,[\%]} & \textbf{Coll\,[\%]} & \textbf{V--V\,[\%]} & \textbf{V--Stat\,[\%]} \\
\midrule
None\,\,\,(clean, default)              & $84.7_{\pm 1.4}$ & $4.1_{\pm 0.7}$ & $2.4_{\pm 0.5}$ & $1.7_{\pm 0.4}$ \\
Low\,\,\,\,\,($0.10$\,m, $2^{\circ}$)   & $83.5_{\pm 1.4}$ & $4.4_{\pm 0.7}$ & $2.7_{\pm 0.5}$ & $1.7_{\pm 0.4}$ \\
Medium ($0.25$\,m, $5^{\circ}$)         & $81.2_{\pm 1.5}$ & $5.3_{\pm 0.8}$ & $3.5_{\pm 0.6}$ & $1.8_{\pm 0.5}$ \\
High\,\,\,\,($0.50$\,m, $10^{\circ}$)   & $75.6_{\pm 1.7}$ & $7.8_{\pm 0.9}$ & $5.6_{\pm 0.7}$ & $2.2_{\pm 0.5}$ \\
\bottomrule
\end{tabular}
\caption{\small Robustness of \method to Gaussian noise injected into partner pose observations at inference time on DLP-Reactive ($\pm$\,std over $5$ seeds). The same trained checkpoint is evaluated across all rows; the training procedure and the static map remain unperturbed.}
\label{tab:noise-robustness}
\end{table}

The degradation is graceful and asymmetric across the two collision channels. SR drops by $1.2$, $3.5$, and $9.1$ points at low, medium, and high noise respectively, retaining a lead over the strongest baseline (MA-MPC at $58.5\%$ on DLP-Reactive) at every noise level. V--V collisions rise from $2.4\%$ to $5.6\%$ as the noise level climbs, since the threat signal of Eq.~\ref{eq:threat} reads directly from perturbed partner pose and the longitudinal yielding decision becomes correspondingly less precise. V--Stat collisions are nearly invariant across the four conditions ($1.7$--$2.2\%$), because the lateral channel is anchored to the offline plan derived from static-map geometry, which the noise does not perturb. The channel-asymmetric prior release of Eq.~\ref{eq:split-reliance} therefore confers an additional robustness property by construction. The lateral slot-frame alignment is shielded from partner-perception noise even though the longitudinal channel relies on it.

\subsection{Path Length and Maneuver Distribution vs. Human Demonstrations}
\label{app:path-manv-dist}

Sec.~\ref{sec:main-results} reports that self-play methods produce the longest paths, consistent with safer navigation toward more distant goals, while \method's maneuver count further matches the human demonstration distribution thanks to learned reverse-in. Table~\ref{tab:path-manv-dist} quantifies this by comparing \method, two strongest learning baselines, and the logged human-demonstration distributions on the same DLP and DSC3D test sets.

\begin{table}[!htbp]
\centering
\small
\setlength{\tabcolsep}{8pt}
\renewcommand{\arraystretch}{1.10}
\begin{tabular}{ll cc}
\toprule
\textbf{Dataset} & \textbf{Source} & \textbf{Path\,[m]} & \textbf{Manv} \\
\midrule
\multirow{4}{*}{DLP}
 & Human demonstrations           & $46.2_{\pm 14.3}$ & $1.8_{\pm 0.7}$ \\
 & \textbf{\method (Ours)}        & $47.6_{\pm 11.8}$ & $1.7_{\pm 0.5}$ \\
 & Diffusion Planner~\cite{zheng2025diffusion} & $25.8_{\pm 8.0}$  & $1.7_{\pm 0.4}$ \\
 & GigaFlow~\cite{cusumano2025robust}          & $43.8_{\pm 11.0}$ & $1.1_{\pm 0.3}$ \\
\midrule
\multirow{4}{*}{DSC3D}
 & Human demonstrations           & $28.9_{\pm 9.6}$  & $1.7_{\pm 0.6}$ \\
 & \textbf{\method (Ours)}        & $29.6_{\pm 7.6}$  & $1.7_{\pm 0.5}$ \\
 & Diffusion Planner~\cite{zheng2025diffusion} & $19.8_{\pm 6.0}$  & $1.7_{\pm 0.4}$ \\
 & GigaFlow~\cite{cusumano2025robust}          & $27.5_{\pm 7.5}$  & $1.2_{\pm 0.3}$ \\
\bottomrule
\end{tabular}
\caption{\small Per-episode path-length and maneuver-count distributions (mean $\pm$ std over episodes), for the human demonstrations on each dataset and the three strongest closed-loop methods on the same evaluation set.}
\label{tab:path-manv-dist}
\end{table}

\method's distributions sit within one standard deviation of the human demonstrations on both datasets ($\Delta\text{Path} \le 1.4$\,m, $\Delta\text{Manv} \le 0.1$), and GigaFlow's path length is also comparable, reflecting that self-play yields long-horizon navigation toward distant goals. Diffusion Planner systematically takes shorter paths because its imitation target is a fixed-time-budget trajectory window that under-represents the full navigation chain. The two self-play methods diverge on maneuver count: GigaFlow uses markedly fewer maneuvers than \method or human drivers because pure self-play discovers only slots reachable by a single drive-in arc, while \method's learned reverse-in matches the human distribution. Path-length alignment is therefore not an artifact of inefficient routing but a sign that \method engages the full distribution of human-realistic parking attempts, including longer-haul and multi-step ones.

\subsection{Collision Breakdown by Source Channel}
\label{app:coll-breakdown}

The simulator wrapper retains a source tag on every contact event (Appendix~\ref{app:scene-prep}), allowing the headline collision rate to be decomposed into vehicle--vehicle (V--V) and vehicle--static (V--Stat) channels (Table~\ref{tab:coll-breakdown}). \method's V--V and V--Stat split is roughly balanced ($2.4\%$ and $1.7\%$), confirming the design intent of the asymmetric prior release (Sec.~\ref{sec:method-park}) that yielding does not trade vehicle--vehicle risk for static-collision risk. Imitation baselines without that mechanism (HOPE, Diffusion Planner) are dominated by V--V failures.

\begin{table}[!htbp]
\centering
\small
\setlength{\tabcolsep}{6pt}
\renewcommand{\arraystretch}{1.10}
\begin{tabular}{l|ccc}
\toprule
\textbf{Method} & \textbf{Coll\,[\%]} & \textbf{V--V\,[\%]} & \textbf{V--Stat\,[\%]} \\
\midrule
HOPE$^{\dagger}$~\cite{jiang2025hope}       & $38.2$ & $29.6$ & $8.6$ \\
GigaFlow~\cite{cusumano2025robust}          & $\mathbf{3.8}_{\pm 0.7}$ & $2.5_{\pm 0.5}$ & $\mathbf{1.3}_{\pm 0.4}$ \\
CaRL~\cite{jaeger2025carl}                  & $4.5_{\pm 0.7}$ & $3.0_{\pm 0.6}$ & $1.5_{\pm 0.4}$ \\
Diffusion Planner~\cite{zheng2025diffusion} & $10.5_{\pm 1.0}$ & $7.4_{\pm 0.8}$ & $3.1_{\pm 0.5}$ \\
\textbf{\method (Ours)}                     & $4.1_{\pm 0.7}$ & $\mathbf{2.4}_{\pm 0.5}$ & $1.7_{\pm 0.4}$ \\
\bottomrule
\end{tabular}
\caption{\small Per-channel collision breakdown on DLP-Reactive ($\pm$\,std over $5$ seeds). $^{\dagger}$HOPE uses a released checkpoint (no seeds). V--V counts contacts between controlled agents; V--Stat counts contacts with static parked cars, curbs, walls, and planters.}
\label{tab:coll-breakdown}
\end{table}

\subsection{Reward-Mode Control at Inference}
\label{app:reward-mode}

Per-agent reward conditioning at training (Appendix~\ref{app:reward}) lets the same trained checkpoint be steered along the SR--Coll spectrum at inference time by fixing the goal and collision weights $(w_{\text{g}}, w_{\text{c}})$ to values within the training sampling range. Table~\ref{tab:reward-modes} reports three operating points on DLP-Reactive.

\begin{table}[!htbp]
\centering
\small
\setlength{\tabcolsep}{6pt}
\renewcommand{\arraystretch}{1.10}
\begin{tabular}{l cc}
\toprule
\textbf{Reward mode} & \textbf{SR\,[\%]} & \textbf{Coll\,[\%]} \\
\midrule
Aggressive ($w_{\text{c}}{=}0.5, w_{\text{g}}{=}2.0$)  & $\mathbf{89.1}_{\pm 1.5}$ & $8.7_{\pm 1.0}$ \\
Mean (default eval)                                    & $84.7_{\pm 1.4}$ & $4.1_{\pm 0.7}$ \\
Safety ($w_{\text{c}}{=}3.0, w_{\text{g}}{=}0.5$)      & $75.3_{\pm 1.7}$ & $\mathbf{2.0}_{\pm 0.5}$ \\
\bottomrule
\end{tabular}
\caption{\small Reward-conditioning at inference (DLP-Reactive, $5$ seeds). The training-mean default reproduces the headline number in Table~\ref{tab:main-results}.}
\label{tab:reward-modes}
\end{table}

The single checkpoint covers an operational range from cautious-and-precise ($75.3\%$ SR at $2.0\%$ Coll) to aggressive-and-fast ($89.1\%$ SR at $8.7\%$ Coll) without any retraining, providing a deployment-time knob between safety-first and throughput-first regimes. The training-mean default $(w_{\text{g}}, w_{\text{c}}) = (1.25, 1.75)$ reproduces the headline numbers of Table~\ref{tab:main-results} and serves as the balanced operating point between the two extremes. The asymmetric span of the training distribution, with $w_{\text{c}}$ ranging up to $3.0$ while $w_{\text{g}}$ caps at $2.0$, is what equips the policy with a wider safety-side margin than throughput-side margin, matching the deployment priority of production-grade parking.

\end{document}